\documentclass[twoside,11pt]{article}

\usepackage{jmlr2e}

\newtheorem{defi}{Definition}

\def\argmin{\mathop{\rm argmin}}

\jmlrheading{14}{2013}{3419-3440}{8/12; Revised 7/13}{11/13}{Wei Sun, Junhui Wang and Yixin Fang}

\ShortHeadings{Consistent Selection of Tuning Parameters via Variable Selection Stability}{Sun, Wang and Fang}
\firstpageno{3419}

\begin{document}

\title{Consistent Selection of Tuning Parameters \\ via Variable Selection Stability}

\author{\name Wei Sun \email sun244@purdue.edu \\
       \addr Department of Statistics\\
       Purdue University\\
       West Lafayette, IN 47907, USA
       \AND
       \name Junhui Wang \email junhui@uic.edu \\
       \addr Department of Mathematics, Statistics, and Computer Science\\
       University of Illinois at Chicago\\
       Chicago, IL 60607, USA
       \AND
       \name Yixin Fang \email Yixin.Fang@nyumc.org \\
       \addr Departments of Population Health and Environmental Medicine \\
       New York University\\
       New York, NY 10016, USA}

\editor{Xiaotong Shen}

\maketitle

\begin{abstract}%
Penalized regression models are popularly used in high-dimensional data analysis to conduct variable selection and model fitting simultaneously. Whereas success has been widely reported in literature, their performances largely depend on the tuning parameters that balance the trade-off between model fitting and model sparsity. Existing tuning criteria mainly follow the route of minimizing the estimated prediction error or maximizing the posterior model probability, such as cross validation, AIC and BIC. This article introduces a general tuning parameter selection criterion based on variable selection stability. The key idea is to select the tuning parameters so that the resultant penalized regression model is stable in variable selection. The asymptotic selection consistency is established for both fixed and diverging dimensions. Its effectiveness is also demonstrated in a variety of simulated examples as well as an application to the prostate cancer data.
\end{abstract}


\section{Introduction}

The rapid advance of technology has led to an increasing demand for modern statistical techniques to analyze data with complex structure such as the high-dimensional data. In high-dimensional data analysis, it is generally believed that only a small number of variables are truly informative while others are redundant. An underfitted model excludes truly informative variables and may lead to severe estimation bias in model fitting, whereas an overfitted model includes the redundant uninformative variables, increases the estimation variance and hinders the model interpretation. Therefore, identifying the truly informative variables is regarded as the primary goal of the high-dimensional data analysis as well as its many real applications such as health studies \citep{Fan06}.

Among other variable selection methods, penalized regression models have been popularly used, which penalize the model fitting with various regularization terms to encourage model sparsity, such as the lasso regression \citep{Tib96}, the smoothly clipped absolute deviation \citep[SCAD,][]{Fan01}, the adaptive lasso \citep{Zou06}, and the truncated $l_1$-norm regression \citep{She12}. In the penalized regression models, tuning parameters are often employed to balance the trade-off between model fitting and model sparsity, which largely affects the numerical performance and the asymptotic behavior of the penalized regression models. For example, \citet*{Zha06} showed that, under the irrepresentable condition, the lasso regression is selection consistent when the tuning parameter converges to 0 at a rate slower than $O(n^{-1/2})$. Analogous results on the choice of tuning parameters have also been established for the SCAD, the adaptive lasso, and the truncated $l_1$-norm regression. Therefore, it is of crucial importance to select the appropriate tuning parameters so that the performance of the penalized regression models can be optimized.

In literature, many classical selection criteria have been applied to the penalized regression models, including cross validation \citep{Sto74}, generalized cross validation \citep{Cra79}, Mallows' $C_p$ \citep{Mal73}, AIC \citep{Aka74} and BIC \citep{Sch78}. Under certain regularity conditions, \citet*{Wan07} and \citet*{Wan09} established the selection consistency of BIC for the SCAD, and \citet*{Zha10} showed the selection consistency of generalized information criterion (GIC) for the SCAD. Most of these criteria follow the route of minimizing the estimated prediction error or maximizing the posterior model probability. To the best of our knowledge, few criteria has been developed directly focusing on the selection of the informative variables.

This article proposes a tuning parameter selection criterion based on variable selection stability. The key idea is that if multiple samples are available from the same distribution, a good variable selection method should yield similar sets of informative variables that do not vary much from one sample to another. The similarity between two informative variable sets is measured by Cohen's kappa coefficient \citep{Coh60}, which adjusts the actual variable selection agreement relative to the possible agreement by chance. Similar stability measures have been studied in the context of cluster analysis \citep{Ben02,Wan10} and variable selection \citep{Mei10}. Whereas the stability selection method \citep{Mei10} also follows the idea of variable selection stability, it mainly focuses on selecting the informative variables as opposed to selecting the tuning parameters for any given variable selection methods. The effectiveness of the proposed selection criterion is demonstrated in a variety of simulated examples and a real application. More importantly, its asymptotic selection consistency is established, showing that the variable selection method with the selected tuning parameter would recover the truly informative variable set with probability tending to one.

The rest of the article is organized as follows. Section 2 briefly reviews the penalized regression models. Section 3 presents the idea of variable selection stability as well as the proposed kappa selection criterion. Section 4 establishes the asymptotic selection consistency of the kappa selection criterion. Simulation studies are given in Section 5, followed by a real application in Section 6. A brief discussion is provided in Section 7, and the Appendix is devoted to the technical proofs.

\section{Penalized Least Squares Regression}

Given that $(\bold{x}_1,y_1),\ldots,(\bold{x}_n,y_n)$ are independent and identically distributed from some unknown joint distribution, we consider the linear regression model
$$
\bold{y}=\bold{X}\beta+\epsilon = \sum_{j=1}^p \beta_j {\bf x}_{(j)} +\epsilon,
$$
where $\beta=(\beta_1,\cdots,\beta_p)^T$, $\bold{y}=(y_1,\cdots,y_n)^T$, $\bold{X}=(\bold{x}_1,\cdots,\bold{x}_n)^T=(\bold{x}_{(1)},\cdots, \bold{x}_{(p)})$ with $\bold{x}_i=$ \linebreak[4] $(x_{i1},\cdots,x_{ip})^T$ or $\bold{x}_{(j)}=(x_{1j},\cdots,x_{nj})^T$, and $\epsilon | \bold{X} \sim N(\bold{0}, \sigma^2\textrm{I}_n)$. When $p$ is large, it is also assumed that only a small number of $\beta_j$'s are nonzero, corresponding to the truly informative variables. In addition, both ${\bf y}$ and ${\bf x}_{(j)}$'s are centered, so the intercept can be omitted in the regression model.

The general framework of the penalized regression models can be formulated as
\begin{equation}
\argmin_{\beta}~\frac{1}{n}\|\bold{y}-\bold{X}\beta\|^2+\sum_{j=1}^{p}{p_{\lambda}(|\beta_j|)},
\label{pls}
\end{equation}
where $\|\cdot\|$ is the Euclidean norm, and $p_{\lambda}(|\beta_j|)$ is a regularization term encouraging sparsity in $\beta$. Widely used regularization terms include the lasso penalty $p_{\lambda}(\theta)=\lambda \theta$ \citep{Tib96}, the SCAD penalty with $p'_{\lambda}(\theta)=\lambda \big(I(\theta\le\lambda)+\frac{(\gamma\lambda-\theta)_{+}}{(\gamma-1)\lambda}I(\theta>\lambda) \big)$ \citep{Fan01}, the adaptive lasso penalty $p_{\lambda}(\theta)=\lambda_j\theta=\lambda\theta/|\hat{\beta}_j|$ \citep{Zou06} with $\hat{\beta}_j$ being some initial estimate of $\beta_j$, and the truncated $l_1$-norm penalty $p_{\lambda}(\theta)=\lambda \min(1,\theta)$ \citep{She12}.

With appropriately chosen $\lambda_n$, all the aforementioned regularization terms have been shown to be selection consistent. Here a penalty term is said to be selection consistent if the probability that the fitted regression model includes only the truly informative variables is tending to one, and $\lambda$ is replaced by $\lambda_n$ to emphasize its dependence on $n$ in quantifying the asymptotic behaviors. In particular, \citet*{Zha06} showed that the lasso regression is selection consistent under the irrepresentable condition when $\sqrt{n}\lambda_n\rightarrow \infty$ and $\lambda_n\rightarrow 0$; \citet*{Fan01} showed that the SCAD is selection consistent when $\sqrt{n} \lambda_n\rightarrow \infty$ and $ \lambda_n\rightarrow 0$; \citet*{Zou06} showed that the adaptive lasso is selection consistent when $n\lambda_n\rightarrow \infty$ and $\sqrt{n}\lambda_n\rightarrow 0$; and \citet*{She12} showed that the truncated $l_1$-norm penalty is also selection consistent when $\lambda_n$ satisfies a relatively more complex constraint.

Although the asymptotic order of $\lambda_n$ is known to assure the selection consistency of the penalized regression models, it remains unclear how to appropriately select $\lambda_n$ in finite sample so that the resultant model in (\ref{pls}) with the selected $\lambda_n$ can achieve superior numerical performance and attain asymptotic selection consistency. Therefore, it is in demand to devise a tuning parameter selection criterion that can be employed by the penalized regression models so that their variable selection performance can be optimized.

\section{Tuning via Variable Selection Stability}

This section introduces the proposed tuning parameter selection criterion based on the concept of variable selection stability. The key idea is that if we repeatedly draw samples from the population and apply the candidate variable selection methods, a desirable method should produce the informative variable set that does not vary much from one sample to another. Clearly, variable selection stability is assumption free and can be used to tune any penalized regression model.

\subsection{Variable Selection Stability}

For simplicity, we denote the training sample as $z^n$. A base variable selection method $\Psi(z^n; \lambda)$ with a given training sample $z^n$ and a tuning parameter $\lambda$ yields a set of selected informative variables ${\cal A} \subset \{1,\cdots,p\}$, called the active set. When $\Psi$ is applied to various training samples, different active sets can be produced. Supposed that two active sets ${\cal A}_1$ and ${\cal A}_2$ are produced, the agreement between ${\cal A}_1$ and ${\cal A}_2$ can be measured by Cohen's kappa coefficient \citep{Coh60},
\begin{equation}
\kappa({\cal A}_1, {\cal A}_2)=\frac{Pr(a)-Pr(e)}{1-Pr(e)}.
\label{kappa}
\end{equation}
Here the relative observed agreement between ${\cal A}_1$ and ${\cal A}_2$ is $Pr(a)=(n_{11}+n_{22})\slash p$, and the hypothetical probability of chance agreement $Pr(e)=(n_{11}+n_{12})(n_{11}+n_{21})\slash p^2+(n_{12}+ n_{22})(n_{21}+n_{22})\slash p^2$, with $n_{11}=|{\cal A}_1 \cap {\cal A}_2|$, $n_{12}=|{\cal A}_1 \cap {\cal A}_2^c|$, $n_{21}=|{\cal A}_1^c \cap {\cal A}_2|$,  $n_{22}=|{\cal A}_1^c \cap {\cal A}_2^c|$, and $|\cdot|$ being the set cardinality. Note that $-1 \leq \kappa({\cal A}_1, {\cal A}_2) \leq 1$, where $\kappa({\cal A}_1, {\cal A}_2)=1$ when ${\cal A}_1$ and ${\cal A}_2$ are in complete agreement with $n_{12}=n_{21}=0$, and $\kappa({\cal A}_1, {\cal A}_2)=-1$ when ${\cal A}_1$ and ${\cal A}_2$ are in complete disagreement with $n_{11}=n_{22}=0$ and $n_{12}=n_{21}=p/2$. For degenerate cases with ${\cal A}_1={\cal A}_2=\emptyset$ or ${\cal A}_1={\cal A}_2=\{1,\ldots,p\}$, we set $\kappa(\emptyset, \emptyset)=\kappa(\{1,\ldots,p\}, \{1,\ldots,p\})=-1$ under the assumption that the true model is sparse and containing at least one informative variable. As a consequence, the kappa coefficient in (\ref{kappa}) is not suitable for evaluating the null model with no informative variable and the complete model with all variables. Based on (\ref{kappa}), the variable selection stability is defined as follows.

\begin{defi}
The variable selection stability of $\Psi(\cdot;\lambda)$ is defined as
\begin{eqnarray*}
s(\Psi,\lambda,n)=E \Big ( \kappa(\Psi(Z_1^n;\lambda),\Psi(Z_2^n;\lambda)) \Big ),
\end{eqnarray*}
where the expectation is taken with respect to $Z_1^n$ and $Z_2^n$, two independent and identically training samples of size $n$, and $\Psi(Z_1^n; \lambda)$ and $\Psi(Z_2^n; \lambda)$ are two active sets obtained by applying $\Psi(\cdot; \lambda)$ to $Z_1^n$ and $Z_2^n$, respectively.
\end{defi}

By definition, $-1 \le s(\Psi,\lambda,n) \le 1$, and large value of $s(\Psi,\lambda,n)$ indicates a stable variable selection method $\Psi(\cdot; \lambda)$. Note that the definition of $s(\Psi,\lambda,n)$ relies on the unknown population distribution, therefore it needs to be estimated based on the only available training sample in practice.

\subsection{Kappa Selection Criterion}

This section proposes an estimation scheme of the variable selection stability based on cross validation, and develops a kappa selection criterion to tune the penalized regression models by maximizing the estimated variable selection stability. Specifically, the training sample $z^n$ is randomly partitioned into two subsets $z_1^m$ and $z_2^m$ with $m=\lfloor n/2 \rfloor$ for simplicity. The base variable selection method $\Psi(\cdot; \lambda)$ is applied to two subsets separately, and then two active sets $\widehat{\cal A}_{1\lambda}$ and $\widehat{\cal A}_{2\lambda}$ are obtained, and $s(\Psi,\lambda,m)$ is estimated as $\kappa(\widehat{\cal A}_{1\lambda},\widehat{\cal A}_{2\lambda})$. Furthermore, in order to reduce the estimation variability due to the splitting randomness, multiple data splitting can be conducted and the average estimated variable selection stability over all splittings is computed. The selected $\lambda$ is then the one obtaining upper $\alpha_n$ quartile of the average estimated variable selection stability. The proposed kappa selection criterion is present as follows.

\indent {\it Algorithm 1 (kappa selection criterion) :}\\
\indent {\it Step 1}. Randomly partition $(\bold{x}_1,\cdots,\bold{x}_n)^T$ into two subsets $z_1^{*b}=(\bold{x}_1^{*b},\cdots,\bold{x}_m^{*b})^T$ and $z_2^{*b}=$ \linebreak[4] $(\bold{x}_{m+1}^{*b},\cdots,\bold{x}_{2m}^{*b})^T$.\\
\indent {\it Step 2}. Obtain $\widehat{\cal A}_{1\lambda}^{*b}$ and $\widehat{\cal A}_{2\lambda}^{*b}$ from $\Psi(z_1^{*b},\lambda)$ and $\Psi(z_2^{*b},\lambda)$ respectively, and estimate the variable selection stability of $\Psi(\cdot; \lambda)$ in the $b$-th splitting by
$$
\hat{s}^{*b}(\Psi,\lambda,m)=\kappa(\widehat{\cal A}_{1\lambda}^{*b}, \widehat{\cal A}_{2\lambda}^{*b}).
$$
\indent {\it Step 3}. Repeat {\it Steps 1-2} for $B$ times. The average estimated variable selection stability of $\Psi(\cdot; \lambda)$ is then
$$
\hat{s}(\Psi,\lambda,m)=B^{-1}\sum_{b=1}^{B} \hat{s}^{*b}(\Psi,\lambda,m).
$$
\indent {\it Step 4}. Compute $\hat{s}(\Psi,\lambda,m)$ for a sequence of $\lambda$'s, and select
$$
\hat{\lambda}=\min \Big\{ \lambda: \frac{\hat{s}(\Psi,\lambda,m)}{\max_{\lambda'} \hat{s}(\Psi,\lambda',m)}  \geq 1-\alpha_n \Big\}.
$$

Note that the treatment in Step $4$ is necessary since some informative variables may have relatively weak effect compared with others. A large value of $\lambda$ may produce an active set that consistently overlooks the weakly informative variables, which leads to an underfitted model with large variable selection stability. To assure the asymptotic selection consistency, the thresholding value $\alpha_n$ in Step $4$ needs to be small and converges to 0 as $n$ grows. Setting $\alpha_n=0.1$ in the numerical experiments yields satisfactory performance based on our limited experience. Furthermore, the sensitivity study in Section \ref{simulation1} suggests that $\alpha_n$ has very little effect on the selection performance when it varies in a certain range. In Steps 1-3, the estimation scheme based on cross-validation can be replaced by other data re-sampling strategies such as bootstrap or random weighting, which do not reduce the sample size in estimating $\widehat{\cal A}_{1\lambda}^{*b}$ and $\widehat{\cal A}_{2\lambda}^{*b}$, but the independence between $\widehat{\cal A}_{1\lambda}^{*b}$ and $\widehat{\cal A}_{2\lambda}^{*b}$ will no longer hold.

The proposed kappa selection criterion shares the similar idea of variable selection stability with the stability selection method \citep{Mei10}, but they differ in a number of ways. First, the stability selection method is a competitive variable selection method, which combines the randomized lasso regression and the bootstrap, and achieves superior variable selection performance. However, the kappa selection criterion can be regarded as a model selection criterion that is designed to select appropriate tuning parameters for any variable selection method. Second, despite of its robustness, the stability selection method still requires a number of tuning parameters. The authors proposed to select the tuning parameters via controlling the expected number of falsely selected variables. However, this criterion is less applicable in practice since the expected number of falsely selected variables can only be upper bounded by an expression involving various unknown quantities. On the contrary, the kappa selection criterion can be directly applied to select the tuning parameters for the stability selection method.

\section{Asymptotic Selection Consistency}

This section presents the asymptotic selection consistency of the proposed kappa selection criterion. Without loss of generality, we assume that only the first $p_0$ variables with $0<p_0<p$ are informative, and denote the truly informative variable set as ${\cal A}_T=\{1,\cdots,p_0\}$ and the uninformative variable set as ${\cal A}_T^c=\{p_0+1,\cdots,p\}$. Furthermore, we denote $r_n \prec s_n$ if $r_n$ converges to 0 at a faster rate than $s_n$, $r_n \sim s_n$ if $r_n$ converges to 0 at the same rate as $s_n$, and $r_n \preceq s_n$ if $r_n$ converges to 0 at a rate not slower than $s_n$.

\subsection{Consistency with Fixed $p$}

To establish the asymptotic selection consistency with fixed $p$, the following technical assumptions are made.

{\it Assumption 1}: There exist positive $r_n$ and $s_n$ such that the base variable selection method is selection consistent if $r_n \prec \lambda_n \prec s_n$. Let $\lambda_n^*$ be such a tuning parameter with $r_n \prec \lambda_n^* \prec s_n$, then $P(\widehat{\cal A}_{\lambda_n^*} = {\cal A}_T) \geq 1-\epsilon_n$ for some $\epsilon_n \rightarrow 0$. In addition, for any positive constant $\lambda_0$, there exists positive $c_0(\lambda_0)$ such that, when $n$ is sufficiently large,
\begin{equation}
P\Big ( \bigcap_{\lambda_0 r_n \le \lambda_n \le \lambda_n^*} \{\widehat{\cal A}_{\lambda_n}={\cal A}_T \} \Big )\ge 1- c_0(\lambda_0),
\label{eqn:assump1}
\end{equation}
where $c_0(\lambda_0)$ converges to $0$ as $\lambda_0\rightarrow \infty$.

Assumption 1 specifies an asymptotic working interval for $\lambda_n$ within which the base variable selection method is selection consistent. Here the consistent rate $\epsilon_n$ is defined for $\lambda_n^*$ only, and needs not hold uniformly over all $\lambda_n$ with $r_n \prec \lambda_n \prec s_n$. Furthermore, (\ref{eqn:assump1}) establishes an uniform lower bound for the probability of selecting the true model when $\lambda_n$ is within the interval $(\lambda_0 r_n, \lambda_n^*)$.

{\it Assumption 2}: Given $r_n$ in Assumption 1, for any positive constant $\lambda_0$, there exist $\zeta_n$, $c_1(\lambda_0)$ and $c_2(\lambda_0)$ such that, when $n$ is sufficiently large,
\begin{eqnarray}
&&\min_{j\in{\cal A}_T}P\Big(\bigcap_{r_n^{-1} \lambda_n \leq \lambda_0} \{j \in \widehat{\cal A}_{\lambda_n}\} \Big ) \ge 1 - \zeta_n,\label{eqn:assump2_1}\\
&&\min_{j\in{\cal A}_T^c}P\Big(\bigcap_{r_n^{-1} \lambda_n \leq \lambda_0} \{j \in \widehat{\cal A}_{\lambda_n}\} \Big ) \ge c_1(\lambda_0),\label{eqn:assump2_2}\\
&&\max_{j\in{\cal A}_T^c}P\Big(\bigcap_{r_n^{-1} \lambda_n \geq \lambda_0} \{j \notin \widehat{\cal A}_{\lambda_n} \} \Big)\ge c_2(\lambda_0),\label{eqn:assump2_3}
\end{eqnarray}
where $\zeta_n \rightarrow 0$ as $n\rightarrow \infty$, $c_1(\lambda_0)$ and $c_2(\lambda_0)$ are positive and only depend on $\lambda_0$, and $c_1(\lambda_0) \rightarrow 1$ as $\lambda_0 \rightarrow 0$.

Assumption 2 implies that if $\lambda_n$ converges to $0$ faster than $r_n$, the base variable selection method will select all the variables asymptotically, and when $\lambda_n$ converges to $0$ at the same rate of $r_n$, the base variable selection method will select any noise variable with an asymptotically positive probability. The inequalities (\ref{eqn:assump2_1})-(\ref{eqn:assump2_3}) also establish uniform lower bounds for various probabilities of selecting informative variables or noise variables.

Assumptions 1 and 2 are mild in that they are satisfied by many popular variable selection methods. For instance, Lemma 2 in the online supplementary material shows that Assumptions 1 and 2 are satisfied by the lasso regression, the adaptive lasso, and the SCAD. The assumptions can also be verified for other methods such as the elastic-net \citep{Zou05}, the adaptive elastic net \citep{Zou09}, the group lasso \citep{Yua06}, and the adaptive group lasso \citep{Wan08}.

Given that the base variable selection method is selection consistent with appropriately selected $\lambda_n$'s, Theorem 1 shows that the proposed kappa selection criterion is able to identify such $\lambda_n$'s.

\begin{theorem}
Under Assumptions 1 and 2, any variable selection method in (\ref{pls}) with $\hat{\lambda}_n$ selected as in Algorithm 1 with $\alpha_n \succ \epsilon_n$ is selection consistent. That is,
$$
\lim_{n \rightarrow \infty}\lim_{B \rightarrow \infty} P(\widehat{\cal A}_{\hat{\lambda}_n}={\cal A}_T) = 1.
$$
\end{theorem}

Theorem 1 claims the asymptotic selection consistency of the proposed kappa selection criterion when $p$ is fixed. That is, with probability tending to one, the selected active set by the resultant variable selection method with tuning parameter $\hat \lambda_n$ contains only the truly informative variables. As long as $\alpha_n$ converges to 0 not too fast, the kappa selection criterion is guaranteed to be consistent. Therefore, the value of $\alpha_n$ is expected to have little effect on the performance of the kappa selection criterion, which agrees with the sensitivity study in Section \ref{simulation1}.

\subsection{Consistency with Diverging $p_n$}

In high-dimensional data analysis, it is of interest to study the asymptotic behavior of the proposed kappa selection criterion with diverging $p_n$, where size of truly informative set $p_{0n}$ may also diverge with $n$. To accommodate the diverging $p_n$ scenario, the technical assumptions are modified as follows.

{\it Assumption 1a}: There exist positive $r_n$ and $s_n$ such that the base variable selection method is selection consistent if $r_n \prec \lambda_n \prec s_n$. Let $\lambda_n^*$ be such a tuning parameter with $r_n \prec \lambda_n^* \prec s_n$, then $P(\widehat{\cal A}_{\lambda_n^*} = {\cal A}_T) \geq 1-\epsilon_n$ for some $\epsilon_n \rightarrow 0$. In addition, for any positive constant $\lambda_0$, there exists positive $c_{0n}(\lambda_0)$ such that, when $n$ is sufficiently large,
\begin{equation}
P\Big ( \bigcap_{\lambda_0 r_n \le \lambda_n \le \lambda_n^*} \{\widehat{\cal A}_{\lambda_n}={\cal A}_T \} \Big )\ge 1- c_{0n}(\lambda_0),
\label{eqn:assump1a}
\end{equation}
where $\lim_{\lambda_0 \rightarrow \infty} \lim_{n \rightarrow \infty} c_{0n}(\lambda_0) \rightarrow 0$.

{\it Assumption 2a}: Given $r_n$ in Assumption 1a, for any positive constant $\lambda_0$, there exist $\zeta_n$, $c_{1n}(\lambda_0)$ and $c_{2n}(\lambda_0)$ such that, when $n$ is sufficiently large,
\begin{eqnarray}
&&\min_{j\in{\cal A}_T}P\Big(\bigcap_{r_n^{-1} \lambda_n \leq \lambda_0} \{j \in \widehat{\cal A}_{\lambda_n}\} \Big ) \ge 1 - \zeta_n,\label{eqn:assump2a_1}\\
&&\min_{j\in{\cal A}_T^c}P\Big(\bigcap_{r_n^{-1} \lambda_n \leq \lambda_0} \{j \in \widehat{\cal A}_{\lambda_n}\} \Big ) \ge c_{1n}(\lambda_0),\label{eqn:assump2a_2}\\
&&\max_{j\in{\cal A}_T^c}P\Big(\bigcap_{r_n^{-1} \lambda_n \geq \lambda_0} \{j \notin \widehat{\cal A}_{\lambda_n} \} \Big)\ge c_{2n}(\lambda_0),\label{eqn:assump2a_3}
\end{eqnarray}
where $\zeta_n$ satisfies $p_n\zeta_n \rightarrow 0$ as $n\rightarrow \infty$, $c_{1n}(\lambda_0)$ and $c_{2n}(\lambda_0)$ are positive and may depend on $n$ and $\lambda_0$, and $\lim_{\lambda_0 \rightarrow 0} \lim_{n \rightarrow \infty} c_{1n}(\lambda_0)=1$.

\begin{theorem}
\label{nonempty2}
Under Assumptions 1a and 2a, any variable selection method in (\ref{pls}) with $\hat{\lambda}_n$ selected as in Algorithm 1 with $\min \big (p_n(1-\tilde{c}_{1n}),  p_n^{-1} c_{1n}c_{2n} \big ) \succ \alpha_n \succ \epsilon_n$ is selection consistent, where $\tilde{c}_{1n}=\sup_{\lambda_0}c_{1n}(\lambda_0)$, $c_{1n}=\inf_{\lambda_0}c_{1n}(\lambda_0)$, and $c_{2n}=\inf_{\lambda_0}c_{2n}(\lambda_0)$.
\end{theorem}

Theorem \ref{nonempty2} shows the asymptotic selection consistency of the proposed kappa selection criterion with satisfied $\alpha_n$ for diverging $p_n$, where the diverging speed of $p_n$ is bounded as in Theorem \ref{nonempty2} and depends on the base variable selection method. Lemma 3 in the online supplementary material shows that (\ref{eqn:assump1a})-(\ref{eqn:assump2a_3}) in Assumptions 1a and 2a are satisfied by the lasso regression. However, it is generally difficult to verify Assumptions 1a and 2a for other  popular variable selection algorithms \citep{Fan04,Hua07,Hua08}, as the convergence rates in both assumptions are not explicitly specified.

\section{Simulations}

This section examines the effectiveness of the proposed kappa selection criterion in simulated examples. Its performance is compared against a number of popular competitors, including Mallows' $C_p$ ($C_p$), BIC, 10-fold cross validation (CV), and generalized cross validation (GCV). Their formulations are given as follows,
\begin{eqnarray}
C_p(\lambda)&=&\frac{SSE_{\lambda}}{\hat \sigma^2 }~-~n~+~2\widehat{df},\label{cp_criterion}\\
BIC(\lambda)&=&\log\Big(\frac{SSE_{\lambda}}{n}\Big)~+~\frac{\log(n)\widehat{df}}{n},\nonumber\\
CV(\lambda)&=&\sum_{s=1}^{10}\sum_{(y_k,x_k)\in T^{-s}}\Big(y_k-\bold{x}_k^T\hat{\beta}^{(s)}(\lambda)\Big)^2, \label{cv}\\
GCV(\lambda)&=&\frac{SSE_{\lambda}}{n(1-\widehat{df}/n)^2},\nonumber
\end{eqnarray}
where $SSE_{\lambda}=\|\bold{y}-\bold{X}\hat{\beta}(\lambda)\|^2$, $\widehat{df}$ is estimated as the number of nonzero variables in $\hat{\beta}(\lambda)$ \citep{Zou07}, and $\hat \sigma^2$ in $(\ref{cp_criterion})$ is estimated based on the saturated model. In $(\ref{cv})$, $T^{s}$ and $T^{-s}$ are the training and validation sets in CV, and $\hat{\beta}^{(s)}(\lambda)$ is the estimated $\beta$ using the training set $T^{s}$ and tuning parameter $\lambda$. The optimal $\hat \lambda$ is then selected as the one that minimizes the corresponding $C_p(\lambda)$, $BIC(\lambda)$, $CV(\lambda)$, or $GCV(\lambda)$, respectively.

To assess the performance of each selection criterion, we report the percentage of selecting the true model over all replicates, as well as the number of correctly selected zeros and incorrectly selected zeros in $\hat{\beta}(\hat \lambda)$. The final estimator $\hat{\beta}(\hat \lambda)$ is obtained by refitting the standard least squares regression based only on the selected informative variables. We then compare the prediction performance through the relative prediction error $RPE=E(\bold{x}^T\hat{\beta}(\hat{\lambda})-\bold{x}^T\beta )^2/\sigma^2$ \citep{Zou06}.

\subsection{Scenario \uppercase\expandafter{\romannumeral1}: Fixed $p$}
\label{simulation1}

The simulated data sets $(\bold{x}_i, y_i)_{i=1}^n$ are generated from the model
$$
y=\bold{x}^T\beta+\sigma\epsilon=\sum_{j=1}^8 \bold{x}_{(j)} \beta_j + \sigma\epsilon,
$$
where $\beta=(3,1.5,0,0,2,0,0,0)^T$, $\sigma=1$, $\bold{x}_{(j)}$ and $\epsilon$ are generated from standard normal distribution, and the correlation between $\bold{x}_{(i)}$ and $\bold{x}_{(j)}$ is set as $0.5^{|i-j|}$. This example has been commonly used in literature, including \citet*{Tib96}, \citet*{Fan01}, and \citet*{Wan07}.

\begin{figure}[htb]
\centering
\begin{small}
\epsfig{file=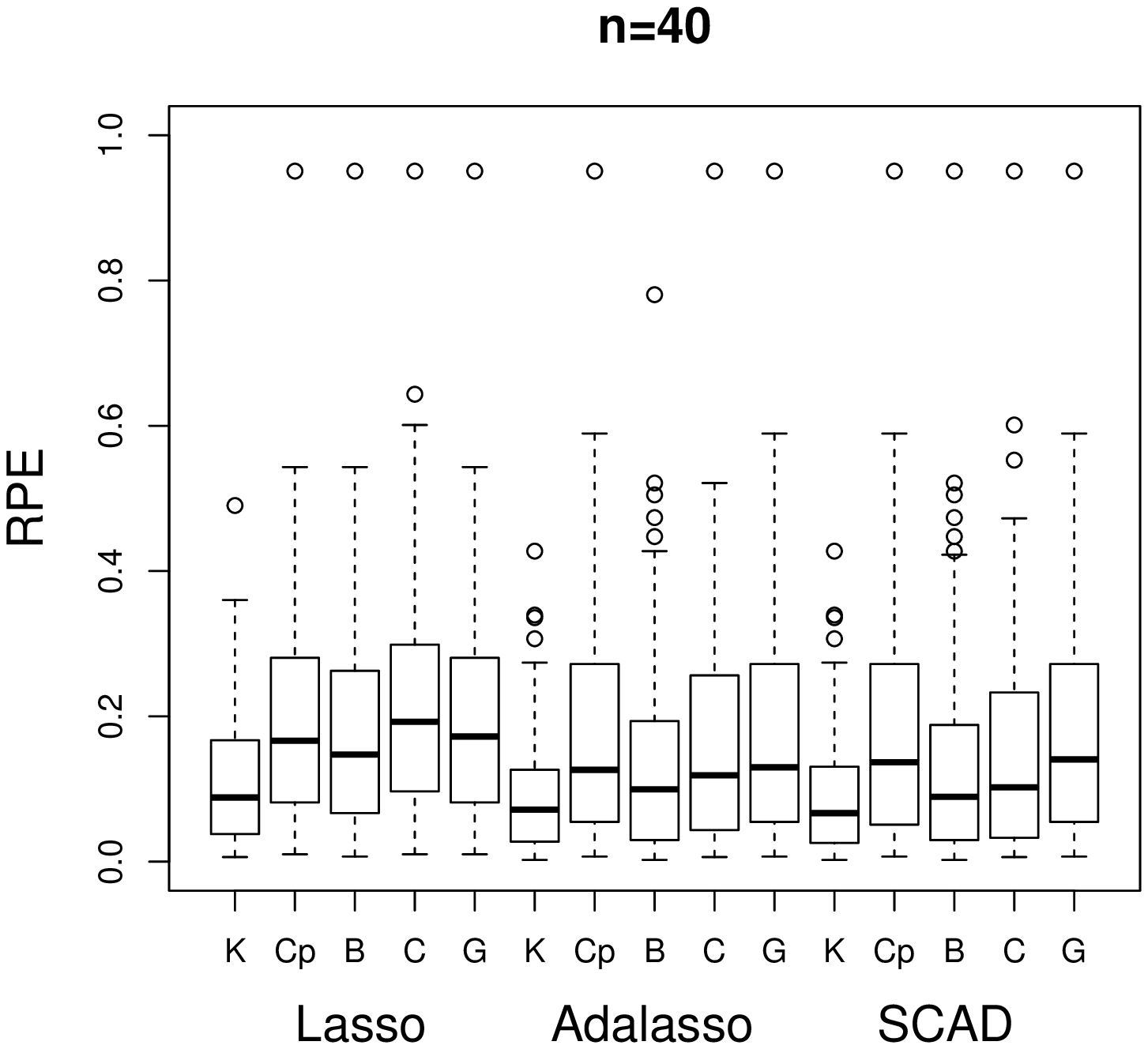, angle=0,width=2.5in}
\epsfig{file=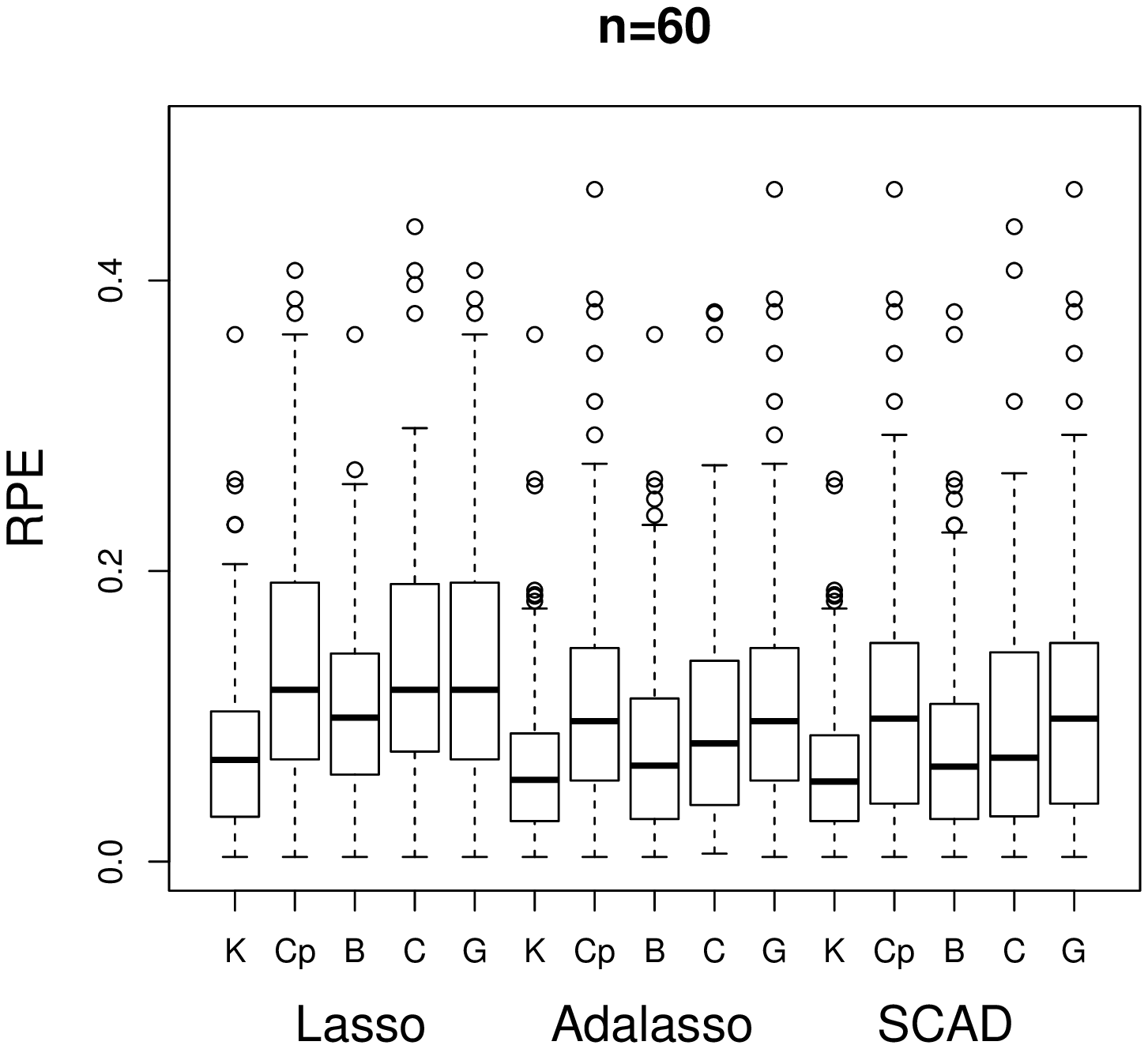, angle=0,width=2.5in}
\epsfig{file=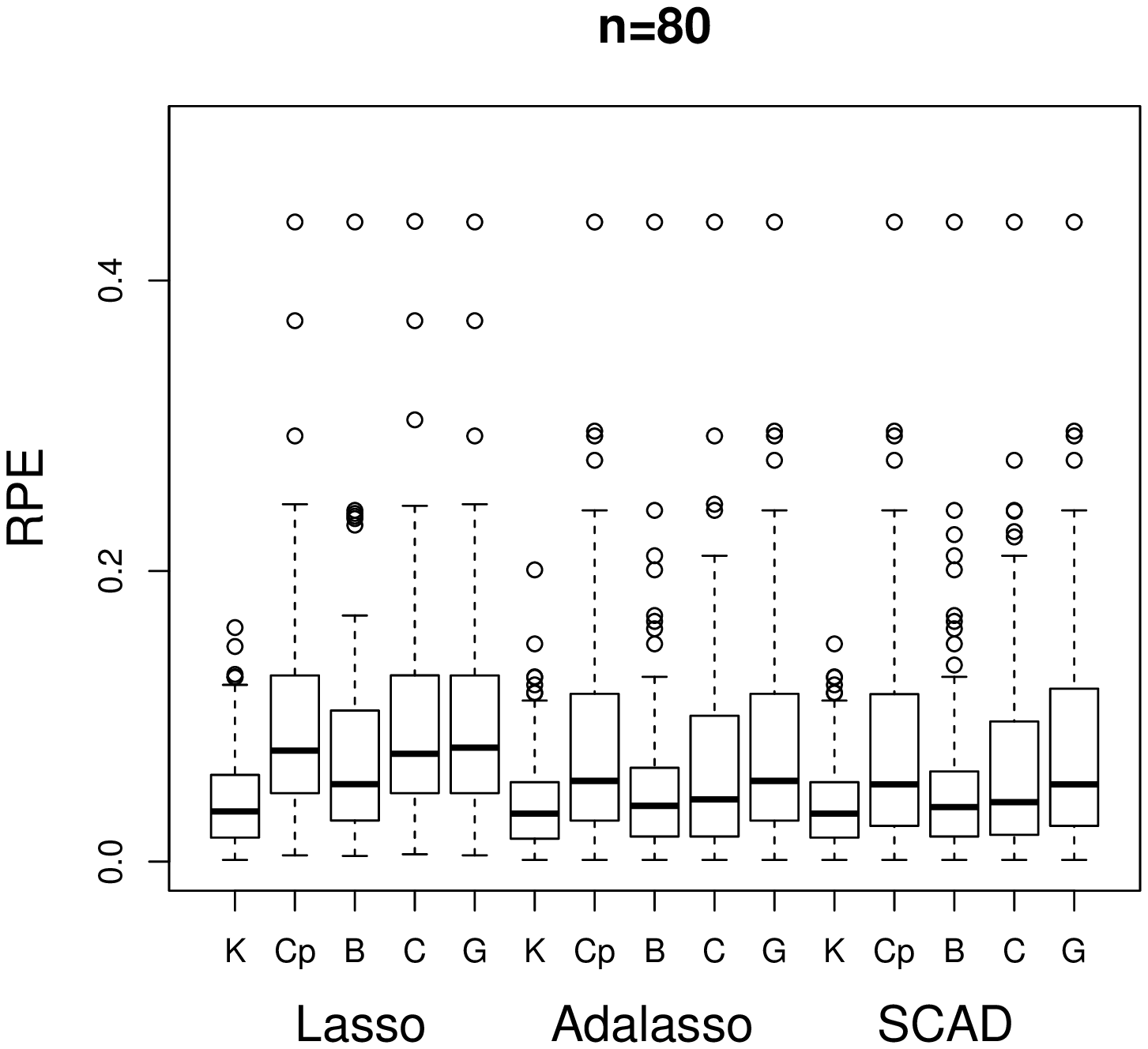, angle=0,width=2.5in}
\end{small}
\caption{ Relative prediction errors (RPE) for various selection criteria in simulations of Section \ref{simulation1}. Here `K', `Cp', `B', `C' and `G' represent the kappa selection criterion, Mallows' $C_p$, BIC, CV and GCV, respectively.}
\label{boxerror1}
\end{figure}

For comparison, we set $n=40$, 60 or 80 and implement the lasso regression, the adaptive lasso and the SCAD as the base variable selection methods. The lasso regression and the adaptive lasso are implemented by package `lars' \citep{Efr04} and the SCAD is implemented by package `ncvreg' \citep{Bre11} in R. The tuning parameter $\lambda$'s are selected via each selection criterion, optimized through a grid search over $100$ points $\{10^{-2+4l/99};~l=0,\ldots,99\}$. The number of splittings $B$ for the kappa selection criterion is $20$. Each simulation is replicated $100$ times, and the percentages of selecting the true active set, the average numbers of correctly selected zeros (C) and incorrectly selected zeros (I), and the relative prediction errors (RPE) are summarized in Tables \ref{per1}-\ref{ci1} and Figure \ref{boxerror1}.

\begin{table}[htb]
\centering
\begin{small}
\begin{tabular}{cr|ccccc}
\hline
$n$ & Penalty~~ & ~Ks~&~Cp~&~BIC~&~CV~&~GCV \\
\hline
& Lasso & \textbf{0.63} & 0.16 & 0.26 & 0.09 & 0.16 \\
40 & Ada lasso & \textbf{0.98} & 0.53 & 0.72 & 0.63 & 0.52 \\
& SCAD & \textbf{0.98} & 0.55 & 0.78 & 0.76 & 0.52 \\
\hline
& Lasso & \textbf{0.81} & 0.16 & 0.32 & 0.14 & 0.17 \\
60 & Ada lasso & \textbf{0.99} & 0.52 & 0.84 & 0.65 & 0.52 \\
& SCAD & \textbf{1} & 0.58 & 0.86 & 0.76 & 0.56 \\
\hline
& Lasso & \textbf{0.89} & 0.16 & 0.38 & 0.08 & 0.16 \\
80 & Ada lasso & \textbf{0.99} & 0.56 & 0.86 & 0.77 & 0.56 \\
& SCAD & \textbf{0.99} & 0.62 & 0.89 & 0.75 & 0.61 \\
\hline
\end{tabular}
\end{small}
\caption{The percentages of selecting the true active set for various selection criteria in simulations of Section \ref{simulation1}. Here `Ks', `Cp', `BIC', `CV' and `GCV' represent the kappa selection criterion, Mallows' $C_p$, BIC, CV and GCV, respectively.}
\label{per1}
\end{table}

\begin{table}[htb]
\centering
\begin{small}
\begin{tabular}{cr|cccccccccc}
\hline
&&~Ks~&~Ks~&~Cp~&~Cp~&~BIC~&~BIC~&~CV~&~CV~&~GCV~&~GCV \\
\hline
$n$ & Penalty~~ & C & I & C & I & C& I & C & I & C & I \\
\hline
& Lasso & \textbf{4.58} & 0.01 & 3.26 & 0 & 3.60 & 0 & 2.66 & 0 & 3.25 & 0\\
40 & Ada lasso & \textbf{4.98} & 0 & 4.16 & 0 & 4.54 & 0 & 4.25 & 0 & 4.15 & 0\\
& SCAD & \textbf{4.99} & 0.01 & 4.11 & 0 & 4.59 & 0 & 4.39 & 0 & 4.06 & 0\\
\hline
& Lasso & \textbf{4.80} & 0 & 3.12 & 0 & 3.91 & 0 & 2.85 & 0 & 3.13 & 0\\
60 & Ada lasso & \textbf{4.99} & 0 & 4.17 & 0 & 4.80 & 0 & 4.35 & 0 & 4.17 & 0\\
& SCAD & \textbf{5} & 0 & 4.15 & 0 & 4.79 & 0 & 4.37 & 0 & 4.12 & 0\\
\hline
& Lasso & \textbf{4.88} & 0 & 3.01 & 0 & 4.02 & 0 & 2.66 & 0 & 3 & 0\\
80 & Ada lasso & \textbf{4.99} & 0 & 4.19 & 0 & 4.80 & 0 & 4.49 & 0 & 4.19 & 0\\
& SCAD & \textbf{4.99} & 0 & 4.23 & 0 & 4.83 & 0 & 4.45 & 0 & 4.22 & 0\\
\hline
\end{tabular}
\end{small}
\caption{The average numbers of correctly selected zeros (C) and incorrectly selected zeros (I) for various selection criteria in simulations of Section \ref{simulation1}. Here `Ks', `Cp', `BIC', `CV' and `GCV' represent the kappa selection criterion, Mallows' $C_p$, BIC, CV and GCV, respectively.}
\label{ci1}
\end{table}

Evidently, the proposed kappa selection criterion delivers superior performance against its competitors in terms of both variable selection accuracy and relative prediction error. As shown in Table \ref{per1}, the kappa selection criterion has the largest probability of choosing the true active set and consistently outperforms other selection criteria, especially when the lasso regression is used as the base variable selection method. As the sample size $n$ increases, the percentage of selecting the true active set is also improving, which supports the selection consistency in Section 4.

Table \ref{ci1} shows that the kappa selection criterion yields the largest number of correctly selected zeros in all scenarios, and it yields almost perfect performance for the adaptive lasso and the SCAD. In addition, all selection criteria barely select any incorrect zeros, whereas the kappa selection criterion is relatively more aggressive in that it has small chance to shrink some informative variables to zeros when sample size is small. All other criteria tend to be conservative and include some uninformative variables, so the numbers of correctly selected zeros are significantly less than $5$.

Besides the superior variable selection performance, the kappa selection criterion also delivers accurate prediction performance and yields small relative prediction error as displayed in Figure \ref{boxerror1}. Note that other criteria, especially $C_p$ and GCV, produce large relative prediction errors, which could be due to their conservative selection of the informative variables.

\begin{figure}[Detection]
\centering
\begin{small}
\epsfig{file=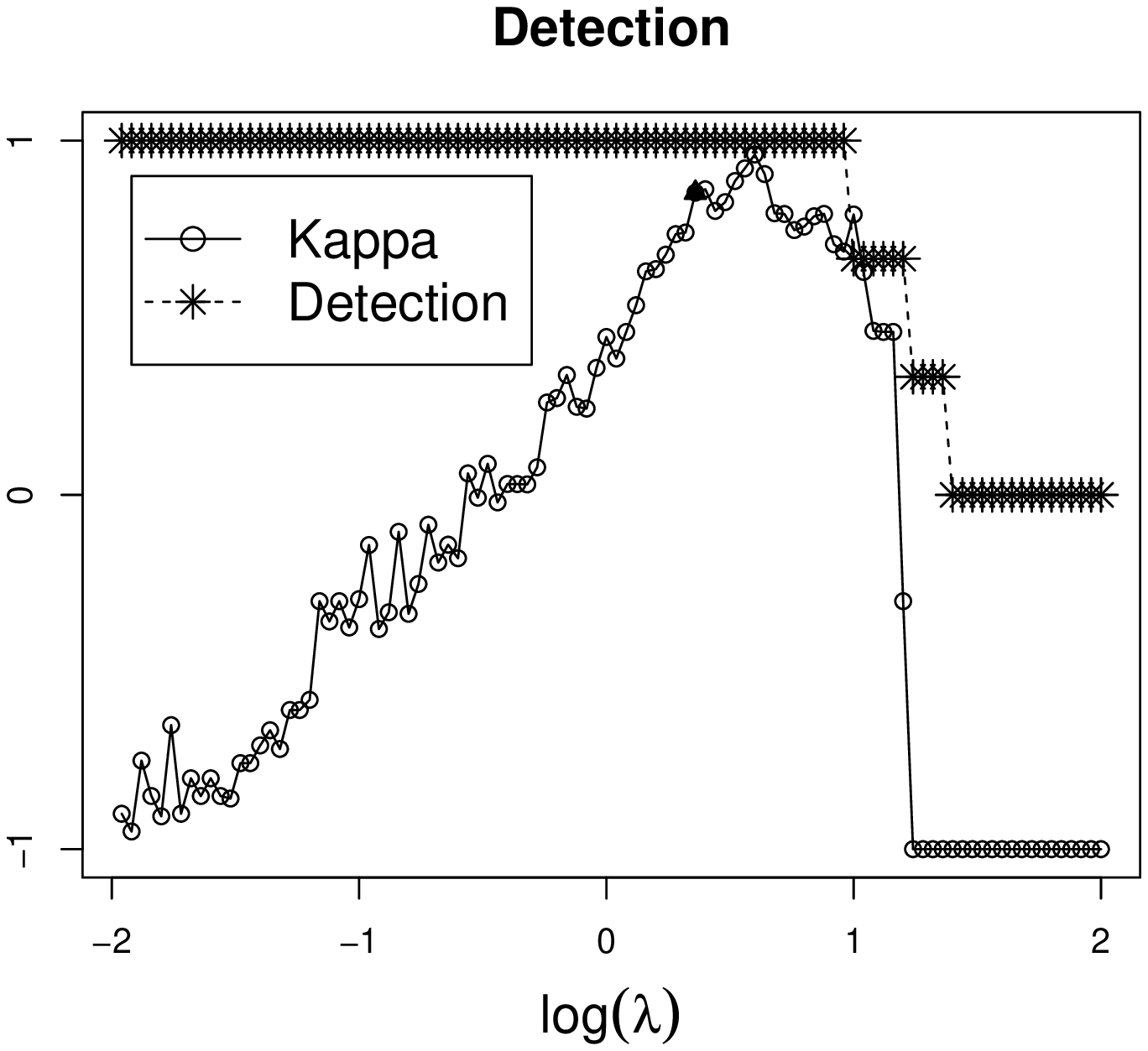, angle=0,width=2.5in}
\epsfig{file=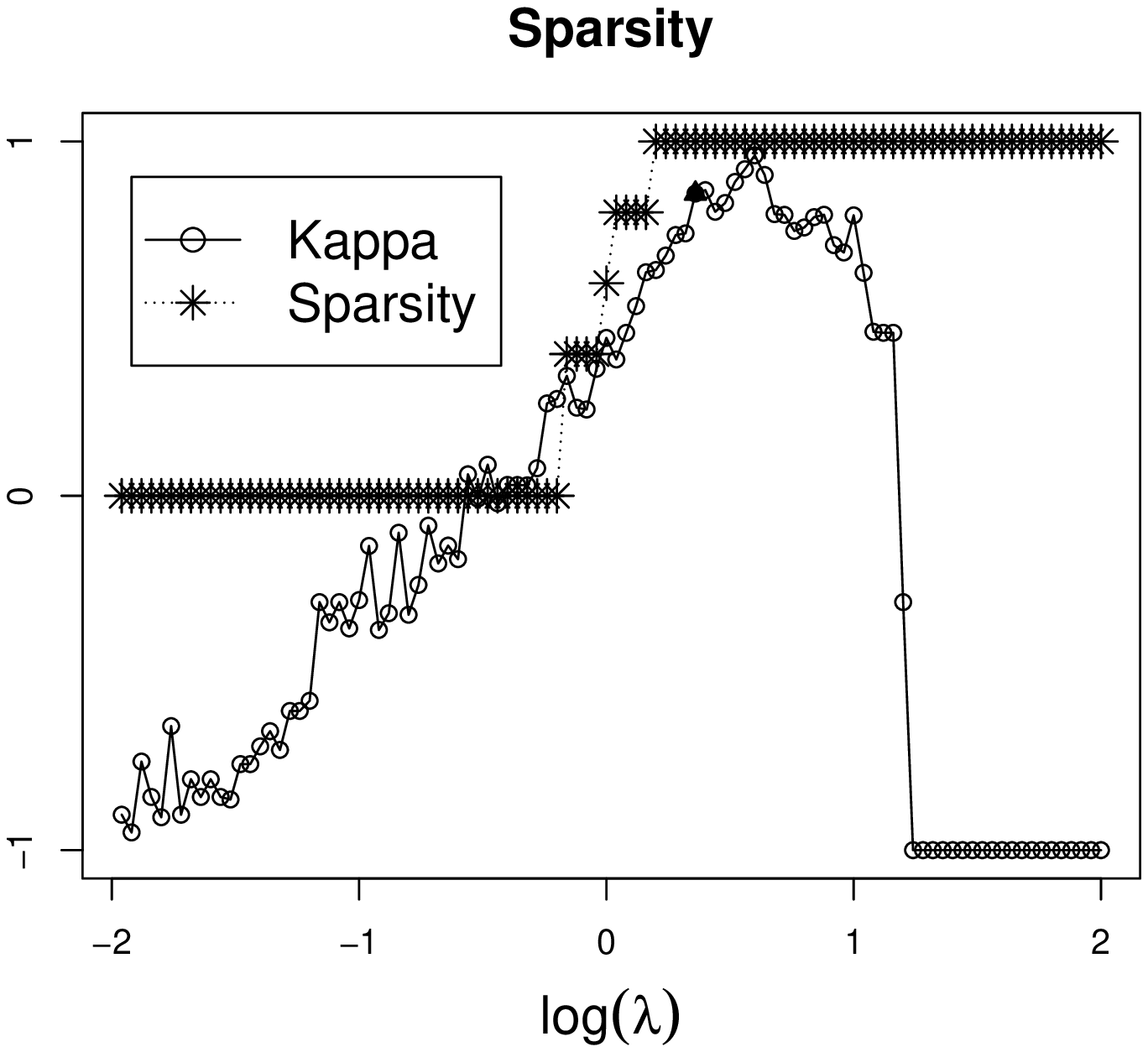, angle=0,width=2.5in}
\epsfig{file=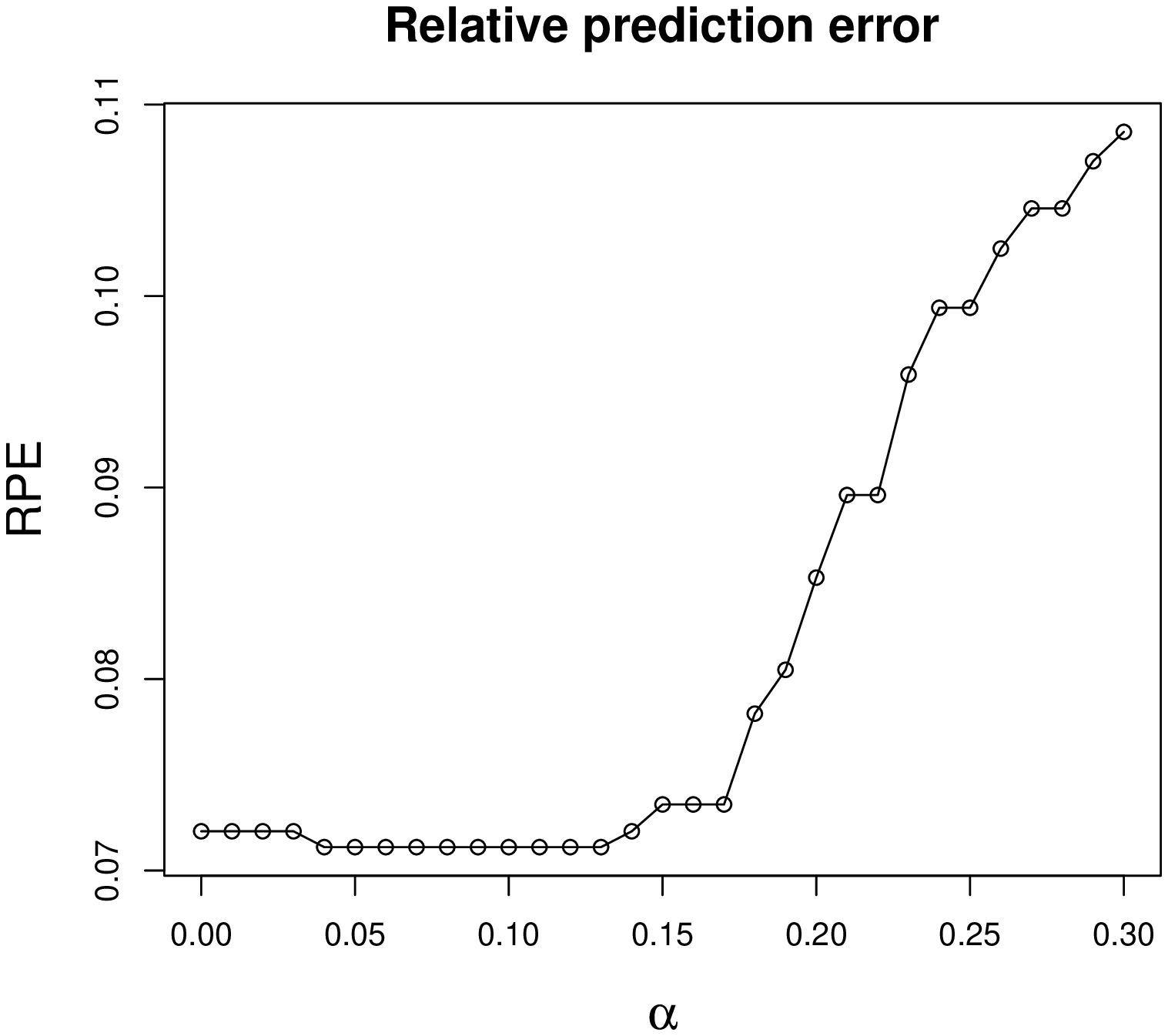, angle=0,width=2.5in}
\end{small}
\caption{The detection and sparsity of the lasso regression with the kappa selection criterion are shown on the top and the sensitivity of $\alpha$ to the relative prediction error is shown on the bottom. The optimal $\log(\lambda)$ selected by the kappa selection criterion is denoted as the filled triangle in the detection and sparsity plots.}
\label{sparsity}
\end{figure}

To illustrate the effectiveness of the kappa selection criterion, we randomly select one replication with $n=40$ and display the estimated variable selection stability as well as the results of detection and sparsity for various $\lambda$'s for the lasso regression. The detection is defined as the percentage of selecting the truly informative variables, and the sparsity is defined as the percentage of excluding the truly uninformative variables. Figure \ref{sparsity} illustrates the clear relationship between the variable selection stability and the values of detection and sparsity. More importantly, the selection performance of the kappa selection criterion is very stable against $\alpha_n$ when it is small. Specifically, we apply the kappa selection criterion on the lasso regression for $\alpha_n =\large\{\frac{l}{100};~ l=0, \ldots, 30\large\}$ and compute the corresponding average RPE over $100$ replications. As shown in the last panel of Figure \ref{sparsity}, the average RPE's are almost the same for $\alpha_n \in (0, 0.13)$, which agrees with the theoretical result in Section 4.

\subsection{Scenario \uppercase\expandafter{\romannumeral2}: Diverging $p_n$}
\label{simulation2}

\begin{figure}[htb]
\centering
\begin{small}
\epsfig{file=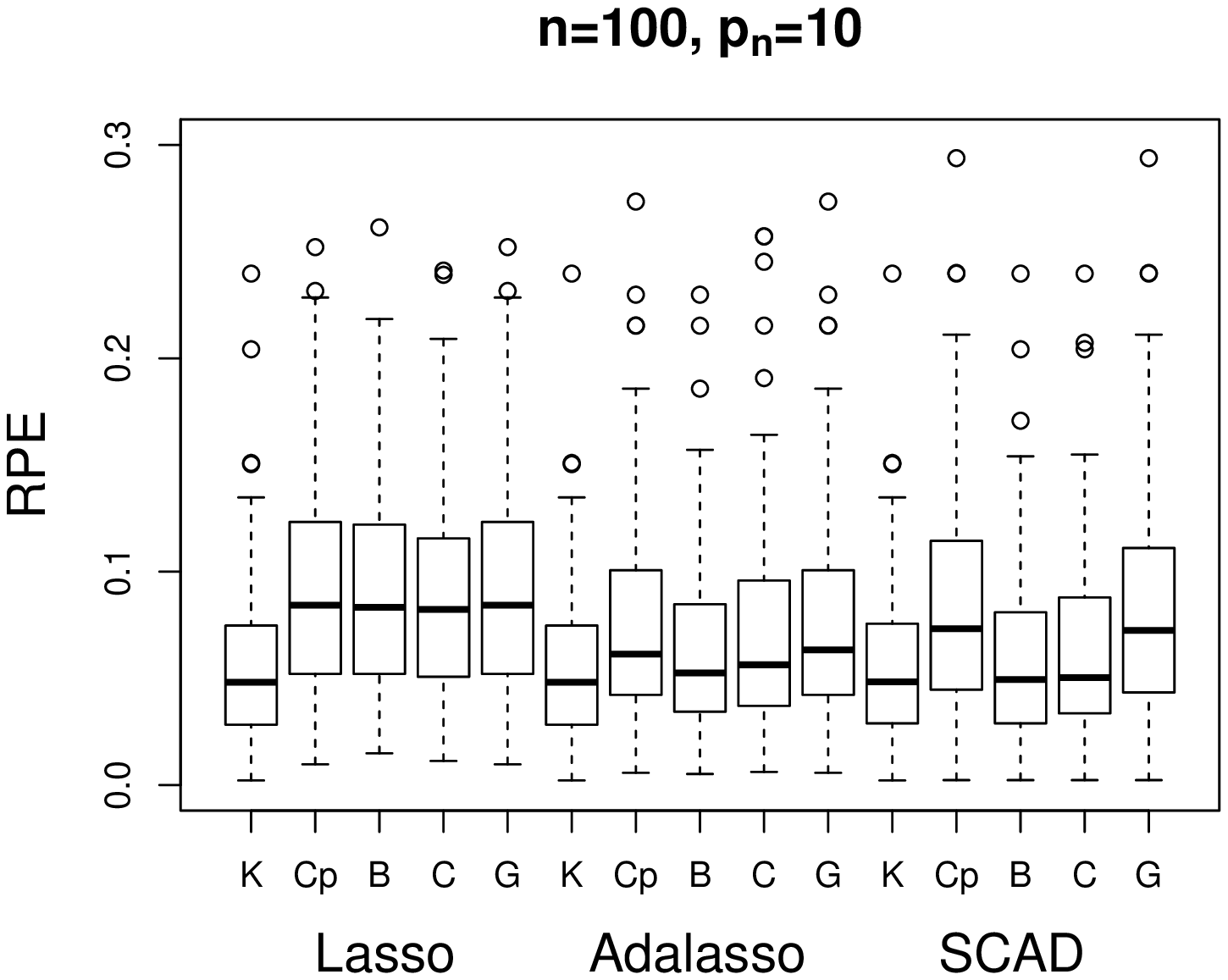, angle=0,width=2.5in}
\epsfig{file=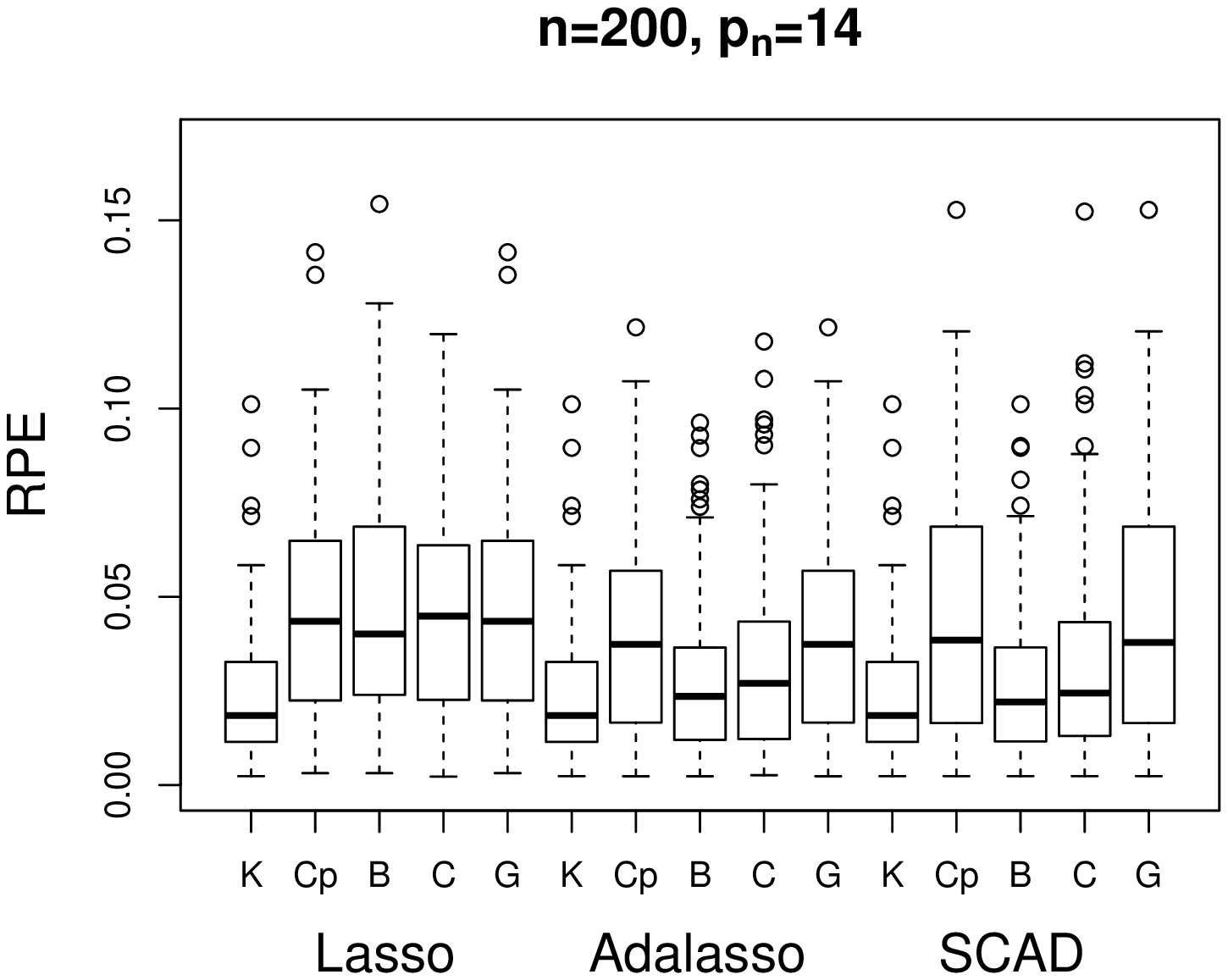, angle=0,width=2.5in}
\epsfig{file=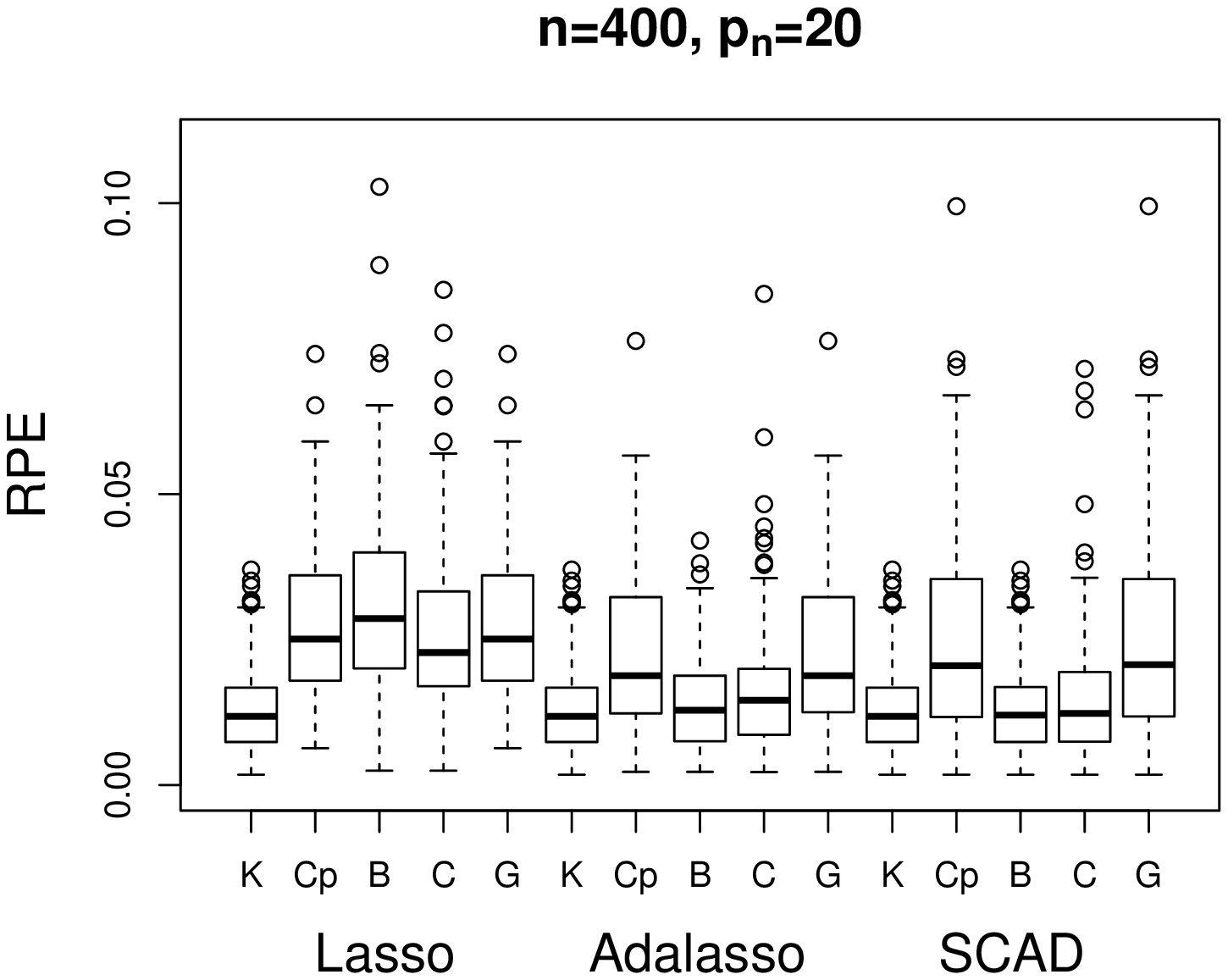, angle=0,width=2.5in}
\epsfig{file=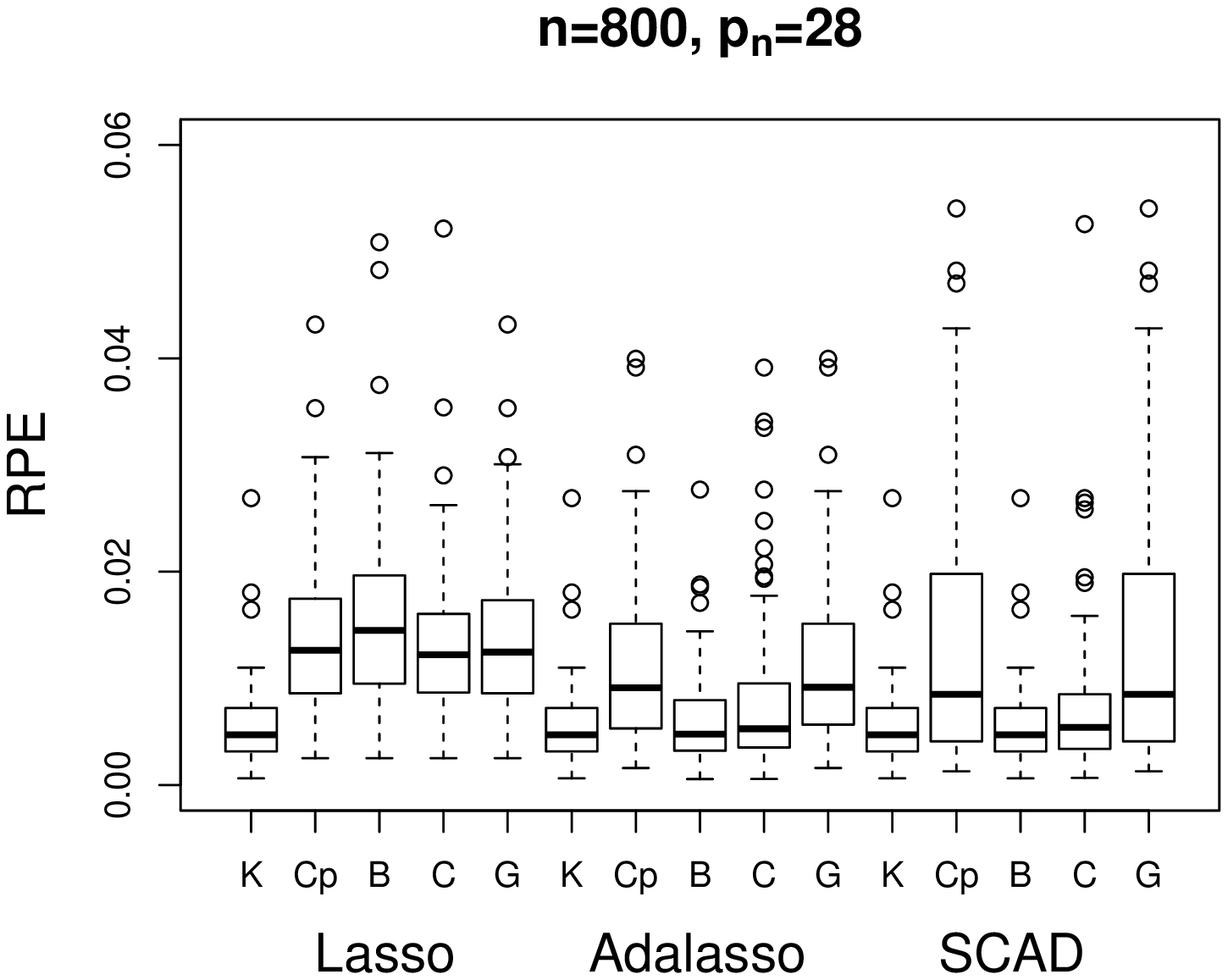, angle=0,width=2.5in}
\end{small}
\caption{ Relative prediction errors (RPE) for various selection criteria in Scenario 2 with $\sigma=1$. Here `K', `Cp', `B', `C' and `G' represent the kappa selection criterion, Mallows' $C_p$, BIC, CV and GCV, respectively.}
\label{boxerrors1}
\end{figure}

\begin{figure}[htb]
\centering
\begin{small}
\epsfig{file=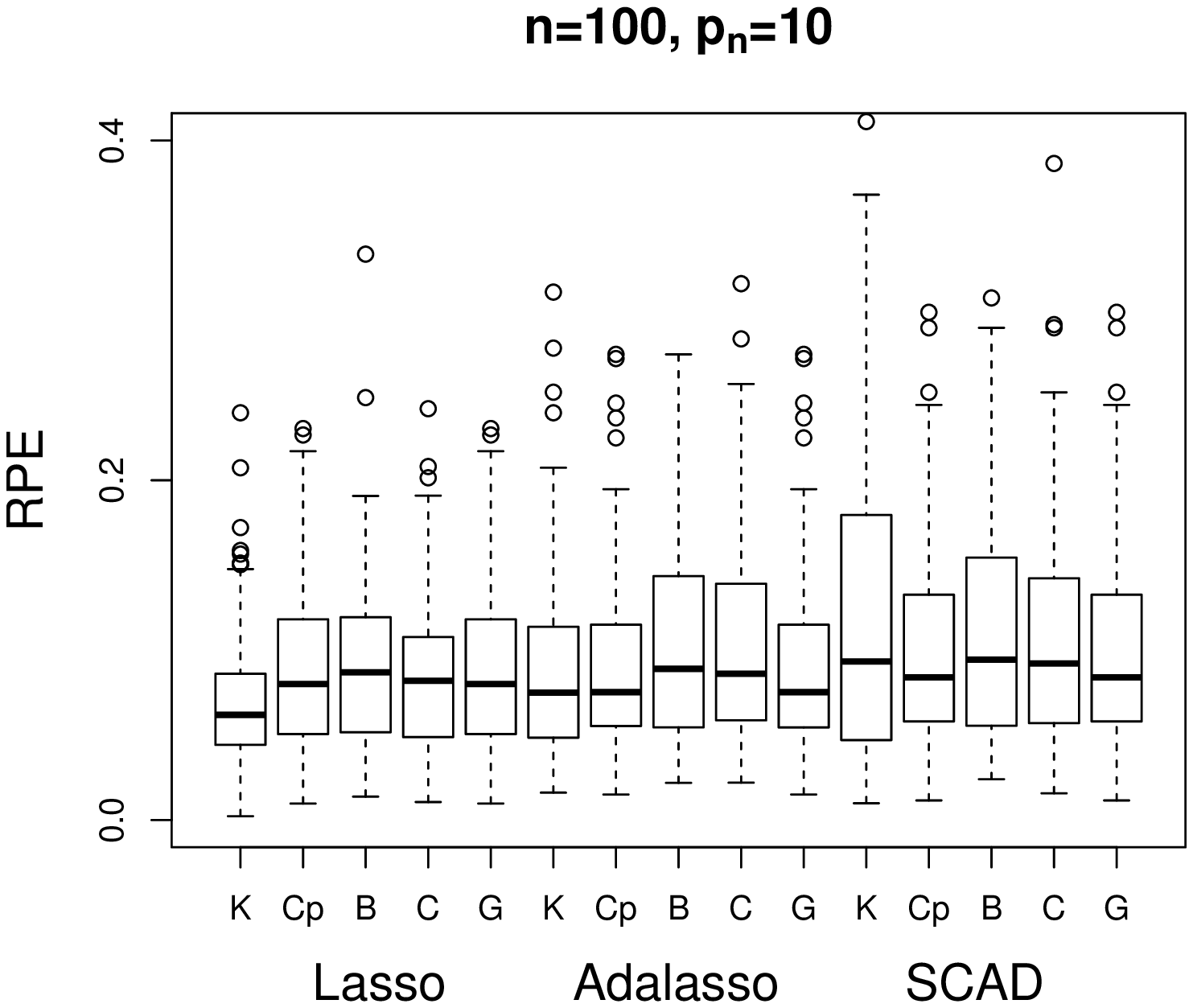, angle=0,width=2.5in}
\epsfig{file=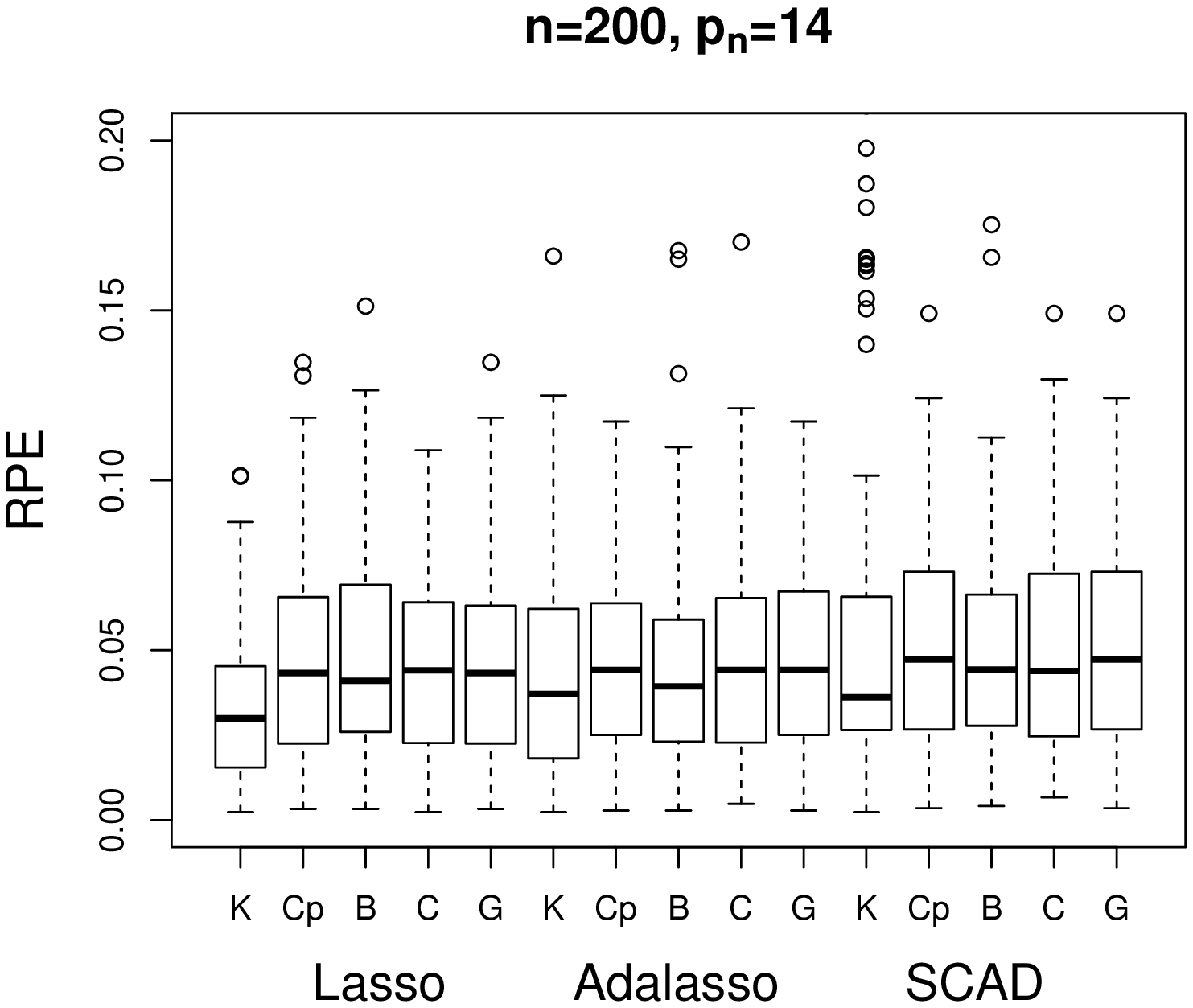, angle=0,width=2.5in}
\epsfig{file=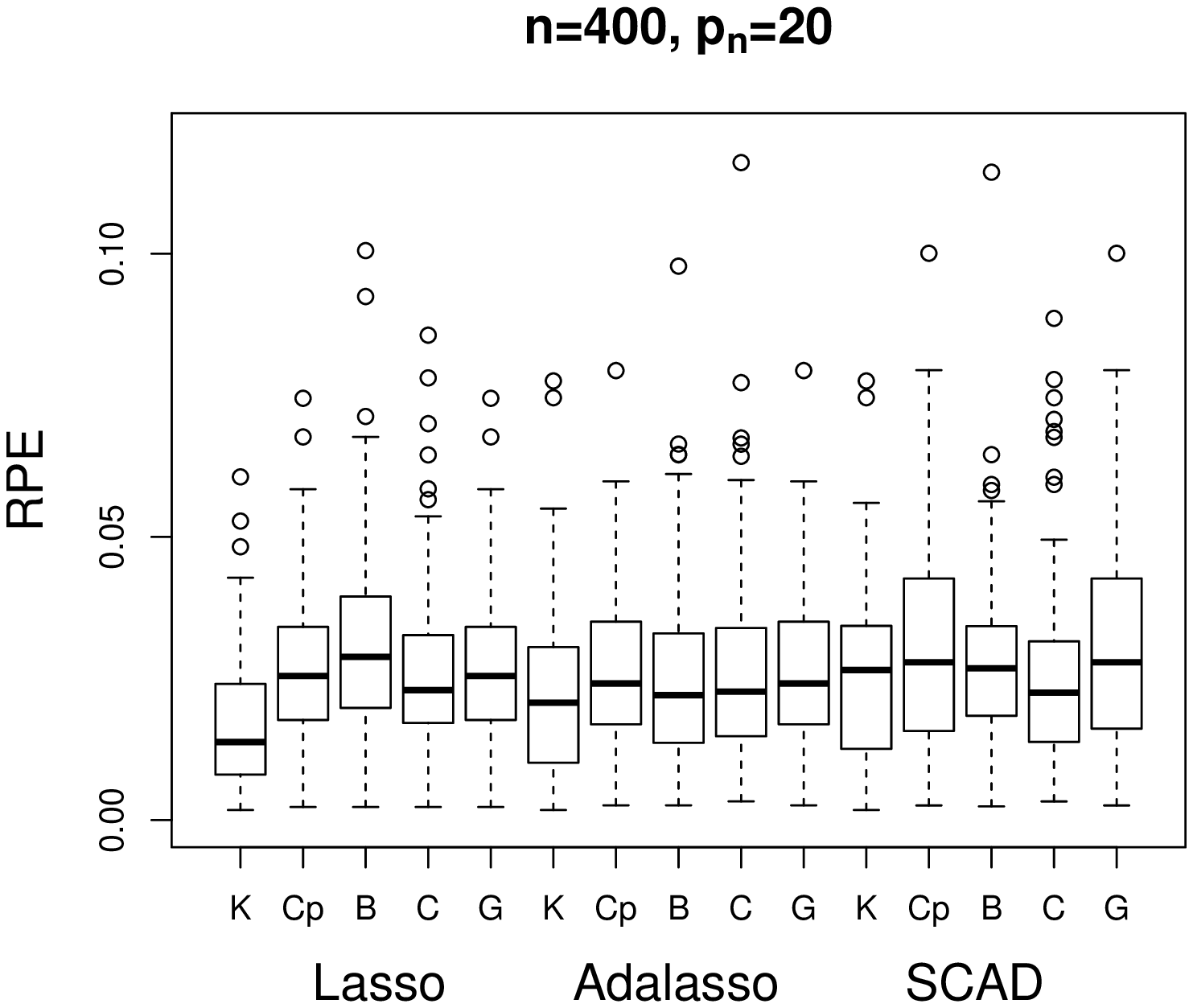, angle=0,width=2.5in}
\epsfig{file=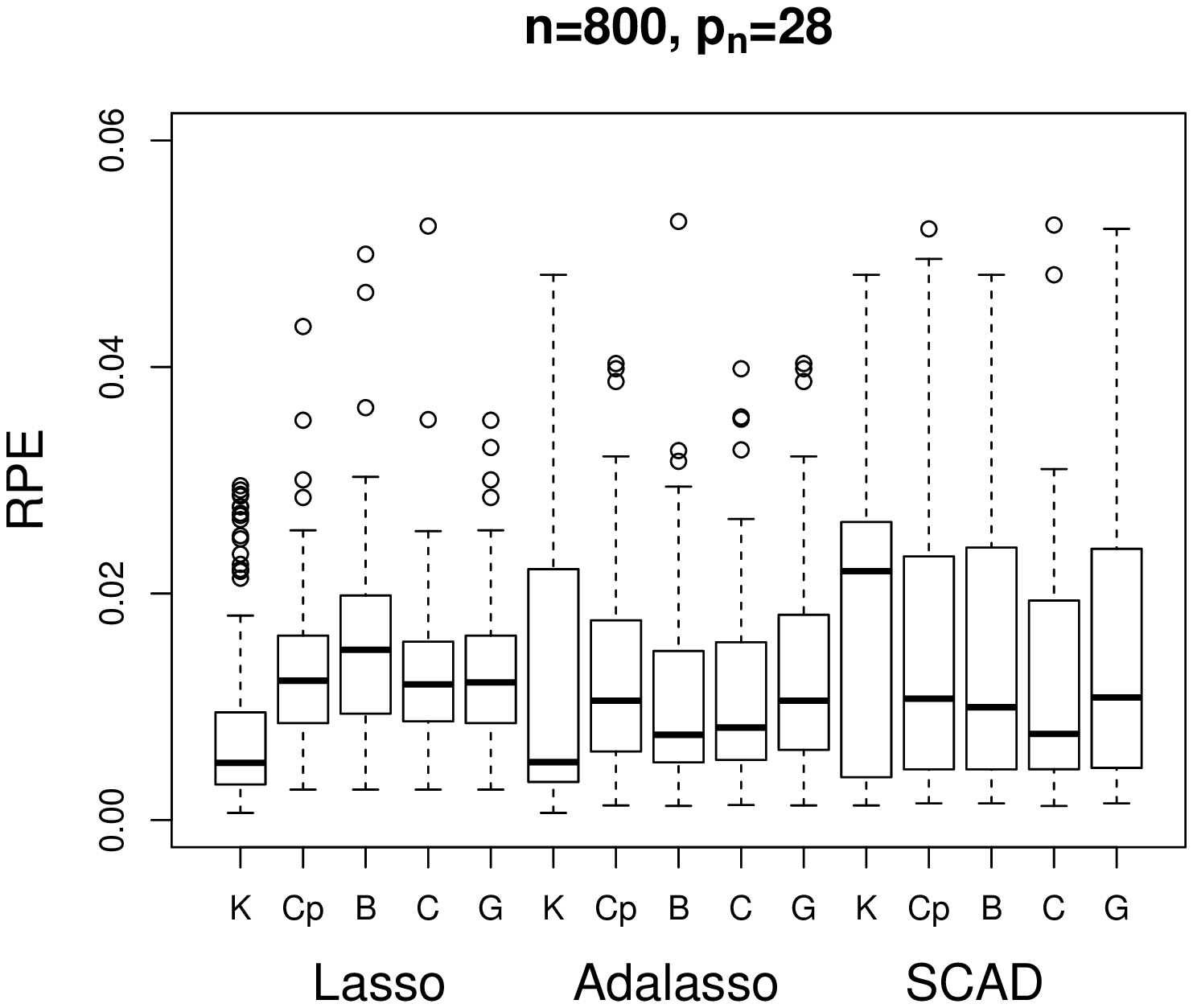, angle=0,width=2.5in}
\end{small}
\caption{ Relative prediction errors (RPE) for various selection criteria in Scenario 2 with $\sigma=6$. Here `K', `Cp', `B', `C' and `G' represent the kappa selection criterion, Mallows' $C_p$, BIC, CV and GCV, respectively.}
\label{boxerrors6}
\end{figure}

To investigate the effects of the noise level and the dimensionality, we compare all the selection criteria in the diverging $p_n$ scenario with a similar simulation model as in Scenario \uppercase\expandafter{\romannumeral1}, except that $\beta=(5,4,3,2,1,0,\cdots,0)^T$, $p_n=[\sqrt{n}]$, and $\sigma=1$ or 6. More specifically, $8$ cases are examined: $n=100,~ p_n=10$; $n=200, ~p_n=14$; $n=400, ~p_n=20$; and $n=800, ~p_n=28$, with $\sigma=1$ or 6 respectively. Note that when $\sigma=6$, the truly informative variables are much more difficult to detect due to the increased noise level. The percentages of selecting the true active set, the average numbers of correctly selected zeros (C) and incorrectly selected zeros (I), and the relative prediction errors (RPE) are summarized in Tables \ref{per2}-\ref{ci2} and Figures \ref{boxerrors1}-\ref{boxerrors6}.

\begin{table}[htb]
\centering
\begin{small}
\begin{tabular}{ccr|ccccc}
\hline
$n$ &$p_n$ &  Penalty~~ & ~Ks~&~Cp~&~BIC~&~CV~&~GCV \\
\hline
&&&\multicolumn{5}{c}{$\sigma=1$} \\
\hline
&& Lasso & \textbf{0.98} & 0.17 & 0.43 & 0.10 & 0.17 \\
100&10 & Ada lasso & \textbf{0.99} & 0.48 & 0.86 & 0.74 & 0.47 \\
&& SCAD & \textbf{0.97} & 0.47 & 0.92 & 0.82 & 0.47 \\
\hline
&& Lasso & \textbf{1} & 0.11 & 0.49 & 0.07 & 0.11 \\
200&14 & Ada lasso & \textbf{1} & 0.38 & 0.90 & 0.66 & 0.38 \\
&& SCAD & \textbf{1} & 0.46 & 0.93 & 0.73 & 0.47 \\
\hline
&& Lasso & \textbf{1} & 0.09 & 0.53 & 0.04 & 0.09 \\
400&20 & Ada lasso & \textbf{1} & 0.34 & 0.93 & 0.73 & 0.33 \\
&& SCAD & \textbf{1} & 0.43 & 0.98 & 0.75 & 0.43 \\
\hline
&& Lasso & \textbf{1} & 0.11 & 0.51 & 0.04 & 0.11 \\
800&28 & Ada lasso & \textbf{1} & 0.30 & 0.96 & 0.74 & 0.29 \\
&& SCAD & \textbf{1} & 0.46 & 0.99 & 0.71 & 0.46 \\
\hline
\hline
&&&\multicolumn{5}{c}{$\sigma=6$} \\
\hline
&& Lasso & \textbf{0.35} & 0.14 & 0.31 & 0.11 & 0.15 \\
100&10 & Ada lasso & \textbf{0.21} & 0.15 & 0.18 & 0.11 & 0.15 \\
&& SCAD & \textbf{0.17} & 0.07 & 0.12 & 0.12 & 0.07 \\
\hline
&& Lasso & \textbf{0.52} & 0.10 & 0.39 & 0.08 & 0.09 \\
200&14 & Ada lasso & \textbf{0.40} & 0.18 & 0.30 & 0.16 & 0.18 \\
&& SCAD & \textbf{0.24} & 0.09 & 0.15 & 0.13 & 0.09 \\
\hline
&& Lasso & \textbf{0.77} & 0.10 & 0.47 & 0.04 & 0.10 \\
400&20 & Ada lasso & 0.53 & 0.22 & \textbf{0.57} & 0.24 & 0.19 \\
&& SCAD & \textbf{0.40} & 0.13 & 0.30 & 0.13 & 0.13 \\
\hline
&& Lasso & \textbf{0.82} & 0.07 & 0.51 & 0.04 & 0.06 \\
800&28 & Ada lasso & \textbf{0.68} & 0.20 & 0.66 & 0.37 & 0.20 \\
&& SCAD & \textbf{0.46} & 0.21 & 0.39 & 0.17 & 0.21 \\
\hline
\end{tabular}
\end{small}
\caption{ The percentages of selecting the true active set for various selection criteria in simulations of Section \ref{simulation2}. Here `Ks', `Cp', `BIC', `CV' and `GCV' represent the kappa selection criterion, Mallows' $C_p$, BIC, CV and GCV, respectively.}
\label{per2}
\end{table}

\begin{table}[htb]
\centering
\begin{small}
\begin{tabular}{ccr|cccccccccc}
\hline
&&&~Ks~&~Ks~&~Cp~&~Cp~&~BIC~&~BIC~&~CV~&~CV~&~GCV~&~GCV \\
\hline
$n$&$p_n$ & Penalty~~ & C & I & C & I & C& I & C & I & C & I \\
\hline
&&&\multicolumn{10}{c}{$\sigma=1$} \\
\hline
&& Lasso & \textbf{5} & 0.02 & 3.25 & 0 & 4.20 & 0 & 2.95 & 0 & 3.25 & 0\\
100&10 & Ada lasso & \textbf{5} & 0.01 & 4.23 & 0 & 4.84 & 0 & 4.48 & 0 & 4.21 & 0\\
&& SCAD & \textbf{5} & 0.03 & 4.12 & 0 & 4.91 & 0 & 4.67 & 0 & 4.15 & 0\\
\hline
&& Lasso & \textbf{9} & 0 & 6.18 & 0 & 8.26 & 0 & 5.62 & 0 & 6.18 & 0\\
200&14 & Ada lasso & \textbf{9} & 0 & 7.50 & 0 & 8.87 & 0 & 8.24 & 0 & 7.50 & 0\\
&& SCAD & \textbf{9} & 0 & 7.43 & 0 & 8.91 & 0 & 8.26 & 0 & 7.47 & 0\\
\hline
&& Lasso & \textbf{15} & 0 & 11.29 & 0 & 14.23 & 0 & 10.56 & 0 & 11.29 & 0\\
400&20 & Ada lasso & \textbf{15} & 0 & 12.93 & 0 & 14.92 & 0 & 14.28 & 0 & 12.91 & 0\\
&& SCAD & \textbf{15} & 0 & 12.67 & 0 & 14.98 & 0 & 14.21 & 0 & 12.64 & 0\\
\hline
&& Lasso & \textbf{23} & 0 & 18.49 & 0 & 22.27 & 0 & 18.20 & 0 & 18.63 & 0\\
800&28 & Ada lasso & \textbf{23} & 0 & 20.31 & 0 & 22.94 & 0 & 22.07 & 0 & 20.23 & 0\\
&& SCAD & \textbf{23} & 0 & 20.21 & 0 & 22.99 & 0 & 21.95 & 0 & 20.21 & 0\\
\hline
\hline
&&&\multicolumn{10}{c}{$\sigma=6$} \\
\hline
&& Lasso & \textbf{4.76} & 0.57 & 3.27 & 0.24 & 4.31 & 0.35 & 3.09 & 0.20 & 3.28 & 0.24\\
100&10 & Ada lasso & 4.57 & 0.77 & 3.81 & 0.54 & \textbf{4.62} & 0.85 & 3.31 & 0.49 & 3.84 & 0.54\\
&& SCAD & \textbf{4.88} & 1.22 & 3.63 & 0.56 & 4.37 & 0.94 & 3.52 & 0.58 & 3.65 & 0.56\\
\hline
&& Lasso & \textbf{8.93} & 0.43 & 6.20 & 0.08 & 8.32 & 0.21 & 5.79 & 0.07 & 6.22 & 0.08\\
200&14 & Ada lasso & \textbf{8.72} & 0.55 & 7.28 & 0.32 & 8.69 & 0.56 & 7.34 & 0.37 & 7.26 & 0.32\\
&& SCAD & \textbf{9} & 0.95 & 7.07 & 0.37 & 8.37 & 0.63 & 7.25 & 0.44 & 7.07 & 0.37\\
\hline
&& Lasso & \textbf{14.98} & 0.21 & 11.46 & 0.03 & 14.21 & 0.07 & 10.60 & 0.03 & 11.45 & 0.03\\
400&20 & Ada lasso & \textbf{14.88} & 0.40 & 12.24 & 0.09 & 14.80 & 0.30 & 12.93 & 0.15 & 12.16 & 0.09\\
&& SCAD & \textbf{15} & 0.67 & 11.97 & 0.13 & 14.65 & 0.51 & 12.66 & 0.23 & 11.88 & 0.12\\
\hline
&& Lasso & \textbf{22.99} & 0.17 & 18.65 & 0.01 & 22.27 & 0.01 & 18.14 & 0.01 & 18.68 & 0.01\\
800&28 & Ada lasso & \textbf{22.96} & 0.29 & 19.84 & 0.02 & 22.71 & 0.16 & 21.19 & 0.04 & 19.71 & 0.02\\
&& SCAD & \textbf{23} & 0.55 & 19.55 & 0.04 & 22.73 & 0.37 & 20.42 & 0.11 & 19.47 & 0.04\\
\hline
\end{tabular}
\end{small}
\caption{ The average numbers of correctly selected zeros (C) and incorrectly selected zeros (I) for various selection criteria in simulations of Section \ref{simulation2}. Here `Ks', `Cp', `BIC', `CV' and `GCV' represent the kappa selection criterion, Mallows' $C_p$, BIC, CV and GCV, respectively.}
\label{ci2}
\end{table}

In the low noise case with $\sigma=1$, the proposed kappa selection criterion outperforms other competitors in both variable selection and prediction performance. As illustrated in Tables \ref{per2}-\ref{ci2}, the kappa selection criterion delivers the largest percentage of selecting the true active set among all the selection criteria, and achieves perfect variable selection performance for all the variable selection methods when $n\ge 200$. Furthermore, as shown in Figure \ref{boxerrors1}, the kappa selection criterion yields the smallest relative prediction error across all cases.

As the noise level increases to $\sigma=6$, the kappa selection criterion still delivers the largest percentage of selecting the true active set among all scenarios except for the adaptive lasso with $n=400$, where the percentage is slightly smaller than that of BIC. As shown in Table \ref{ci2}, the kappa selection criterion yields the largest number of correctly selection zeros. However, it has relatively higher chance of shrinking the fifth informative variable to zero, while the chance is diminishing as $n$ increases.  This phenomenon is also present for BIC. Considering the smaller relative prediction errors achieved by the kappa selection criterion and BIC, these two criteria tend to produce sparser models with satisfactory prediction performance. In practice, if false negatives are of concern, one can increase the thresholding value $\alpha_n$ in the kappa selection criterion, to allow higher tolerance of instability and hence decrease the chance of claiming false negatives. In addition, as shown in Figure \ref{boxerrors6}, the kappa selection criterion yields the smallest relative prediction error for the lasso regression and the adaptive lasso among all scenarios, whereas the advantage is considerably less significant for the SCAD. This is somewhat expected as the SCAD is sensitive to the noise level \citep{Zou06}, which may lead to inaccurate estimation of the variable selection stability.

\section{Real Application}

In this section, we apply the kappa selection criterion to the prostate cancer data \citep{Sta89}, which were used to study the relationship between the level of log(prostate specific antigen) ($lpsa$) and a number of clinical measures. The data set consisted of $97$ patients who had received a radical prostatectomy, and eight clinical measures were log(cancer volume) ($lcavol$), log(prostate weight) ($lweight$), $age$, log(benign prostaic hyperplasia amount) ($lbph$), seminal vesicle invasion ($svi$), log(capsular penetration) ($lcp$), Gleason score ($gleason$) and percentage Gleason scores 4 or 5 ($pgg45$).

The data set is randomly split into two halves: a training set with 67 patients and a test set with 30 patients. Similarly as in the simulated examples, the tuning parameter $\lambda$'s are selected through a grid search over $100$ grid points $\{10^{-2+4l/99};~l=0,\ldots,99\}$ on the training set. Since it is unknown whether the clinical measures are truly informative or not, the performance of all the selection criteria are compared by computing their corresponding prediction errors on the test set in Table \ref{real}.

\begin{table}[htb]
\centering
\begin{small}
\begin{tabular}{cr|ccccc}
\hline
&Penalty~~&~Ks~&~Cp~&~BIC~&~CV~&~GCV~ \\
\hline
Active&Lasso & 1,2,4,5 & 1,2,3,4,5,6,7,8 & 1,2,4,5 & 1,2,3,4,5,7,8 & 1,2,3,4,5,6,7,8 \\
Set&Ada lasso & 1,2,5 & 1,2,3,4,5  & 1,2,3,4,5 & 1,2,3,4,5,6,7,8  & 1,2,3,4,5 \\
&SCAD & 1,2,4,5 & 1,2,3,4,5 & 1,2,3,4,5 & 1,2,3,4,5,6,7,8 & 1,2,3,4,5\\
\hline
& Lasso & \textbf{0.734} & 0.797 & \textbf{0.734} & 0.807 & 0.797\\
PE& Ada lasso & 0.806 & 0.825 & 0.825 & \textbf{0.797} & 0.825\\
& SCAD & \textbf{0.734} & 0.825 & 0.825 & 0.797 & 0.825\\
\hline
\end{tabular}
\end{small}
\caption{ The selected active sets and the prediction errors (PE) for various selection criteria in the prostate cancer example. Here `Ks', `Cp', `BIC', `CV' and `GCV' represent the kappa selection criterion, Mallows' $C_p$, BIC, CV and GCV, respectively.}
\label{real}
\end{table}

As shown in Table \ref{real}, the proposed kappa selection criterion yields the sparsest model and achieves the smallest prediction error for the lasso regression and the SCAD, while the prediction error for the adaptive lasso is comparable to the minima. Specifically, the lasso regression and the SCAD with the kappa selection criterion include $lcavol$, $lweight$, $lbph$ and $svi$ as the informative variables, and the adaptive lasso with the kappa selection criterion selects only $lcavol$, $lweight$ and $svi$ as the informative variables. As opposed to the sparse regression models produced by other selection criteria, the variable $age$ is excluded by the kappa selection criterion for all base variable selection methods, which agrees with the findings in \citet*{Zou05}.

\section{Discussion}

This article proposes a tuning parameter selection criterion based on the concept of variable selection stability. Its key idea is to select the tuning parameter so that the resultant variable selection method is stable in selecting the informative variables. The proposed criterion delivers superior numerical performance in a variety of experiments. Its asymptotic selection consistency is also established for both fixed and diverging dimensions. Furthermore, it is worth pointing out that the idea of stability is general and can be naturally extended to a broader framework of model selection, such as the penalized nonparametric regression \citep{Xue10} and the penalized clustering \citep{Sun12}.

\section{Supplementary Material}
Lemmas $2$ and $3$ and their proofs are provided as online supplementary material for this article.

\acks{The authors thank the Action Editor and three referees for their constructive comments and suggestions, which have led to a significantly improved paper.}

\appendix
\section{}

The lemma stated below shows that if a variable selection method is selection consistent and $\epsilon_n \prec \alpha_n$, then its variable selection stability converges to 1 in probability.

\begin{lemma}
\label{nonempty}
Let $\lambda_n^*$ be as defined in Assumption 1. For any $\alpha_n$,
$$
P\Big(\hat{s}(\Psi,\lambda_n^*,m)\ge 1-\alpha_n\Big)\ge 1- 2\epsilon_n/\alpha_n.
$$
\end{lemma}

\noindent {\bf Proof of Lemma \ref{nonempty}:}
We denote $\widehat{\cal A}_{1\lambda_n^*}^{*b}$ and $\widehat{\cal A}_{2\lambda_n^*}^{*b}$ as the corresponding active sets obtained from two sub-samples at the $b$-th random splitting. Then the estimated variable selection stability based on the $b$-th splitting can be bounded as
$$
P\Big(\hat{s}^{*b}(\Psi,\lambda_n^*,m)=1\Big)= P\Big(\widehat{\cal A}_{1\lambda_n^*}^{*b}=\widehat{\cal A}_{2\lambda_n^*}^{*b}\Big) \ge  P\Big(\widehat{\cal A}_{1\lambda_n^*}^{*b}={\cal A}_T\Big)^2 \ge (1-\epsilon_n)^2 \ge 1-2\epsilon_n.
$$
By the fact that $0\le \hat{s}^{*b}(\Psi,\lambda_n^*,n)\le 1$, we have
$$
E\Big(\hat{s}(\Psi,\lambda_n^*,m)\Big)=E\Big(B^{-1}\sum_{b=1}^{B}\hat{s}^{*b}(\Psi,\lambda_n^*,m)\Big)\ge 1-2\epsilon_n,
$$
and $0\le \hat{s}(\Psi,\lambda_n^*,n)\le 1$. Finally, Markov inequality yields that
$$
P\Big(1-\hat{s}(\Psi,\lambda_n^*,m)\ge \alpha_n\Big)\le \frac{E\Big(1-\hat{s}(\Psi,\lambda_n^*,m)\Big)}{\alpha_n}\le\frac{2\epsilon_n}{\alpha_n},
$$
which implies the desired result immediately.  \hfill $\blacksquare$\\

\noindent{\bf Proof of Theorem 1:} We first show that for any $\epsilon>0$,
$$
\lim_{n \rightarrow \infty} P \Big (\hat \lambda_n > \lambda_n^*~\mbox{or}~ r_n^{-1} \hat \lambda_n \leq 1/\epsilon \Big ) = 0.
$$
Denote $\Omega_1=\{\lambda_n: \lambda_n > \lambda_n^* \}$, $\Omega_2=\{\lambda_n: r_n^{-1} \lambda_n \le \tau \}$ and $\Omega_3=\{\lambda_n: \tau \leq r_n^{-1} \lambda_n \leq 1/\epsilon \}$, where $\tau<1/\epsilon$, $c_1(\tau) \geq 1-1/p$. It then suffices to show that for any $\epsilon>0$,
$$
P \Big ( \hat \lambda_n  \in \Omega_1 \cup \Omega_2 \cup \Omega_3 \Big ) \rightarrow 0.
$$

First, the definition of $\hat{\lambda}_n$ and Lemma 1 imply that
$$
P(\hat{\lambda}_n\le \lambda_n^*) \ge P\Big(\frac{\hat{s}(\Psi,\lambda_n^*,m)}{\max_{\lambda} \hat{s}(\Psi,\lambda,m)}\ge 1-\alpha_n\Big)\ge P\Big(\hat{s}(\Psi,\lambda_n^*,m)\ge 1-\alpha_n\Big) \ge 1- \frac{2\epsilon_n}{\alpha_n}.
$$
This, together with $\epsilon_n \prec \alpha_n$, yields that
$$
P \Big ( \hat \lambda_n  \in \Omega_1 \Big ) = P(\hat{\lambda}_n > \lambda_n^*) \leq \frac{2\epsilon_n}{\alpha_n} \rightarrow 0.
$$

Next, the definition of $\hat \lambda_n$ implies that
$$
\hat{s}(\Psi,\hat \lambda_n,m) \geq (1-\alpha_n) \max_{\lambda} \hat{s}(\Psi,\lambda,m) \geq (1-\alpha_n) \hat{s}(\Psi,\lambda_n^*,m).
$$
This, together with Lemma 1, leads to
\begin{eqnarray}
P \Big ( \hat{s}(\Psi,\hat \lambda_n,m) \geq 1-2\alpha_n \Big ) &\ge& P \Big ( \hat{s}(\Psi,\hat \lambda_n,m) \geq (1-\alpha_n)^2 \Big )\nonumber\\
&\ge& P\Big(\hat{s}(\Psi,\lambda_n^*,m)\ge 1-\alpha_n\Big) \ge 1- \frac{2\epsilon_n}{\alpha_n},\nonumber
\end{eqnarray}
and hence when $\epsilon_n \prec \alpha_n$,
$$
P \Big ( \hat{s}(\Psi,\hat \lambda_n,m) \geq 1-2\alpha_n \Big ) \rightarrow  1.
$$
Therefore, to show $P ( \hat \lambda_n  \in \Omega_2 ) \rightarrow 0$, it suffices to show
\begin{equation}
P \Big (\sup_{\lambda_n \in \Omega_2}  \hat{s}(\Psi,\lambda_n,m) < 1 - 2\alpha_n \Big ) \rightarrow 1. \label{toshow0}
\end{equation}
But Assumption 2 implies that for any $j \in {\cal A}_T^c$ and $j_1 \in {\cal A}_T$, we have
$$
P\Big(j \in \bigcap_{ \lambda_n \in \Omega_2} \widehat{\cal A}_{\lambda_n} \Big ) \ge c_1(\tau) \ge 1-\frac{1}{p} ~~\textrm{and}~~
P\Big(j_1 \in \bigcap_{ \lambda_n \in \Omega_2} \widehat{\cal A}_{\lambda_n} \Big ) \ge 1-\zeta_n.
$$
It implies that
\begin{eqnarray}
&&\lim_{n\rightarrow \infty}P\Big(\{1,\ldots,p\} \in \bigcap_{ \lambda_n \in \Omega_2} \widehat{\cal A}_{\lambda_n} \Big ) \nonumber\\
&\ge& \lim_{n\rightarrow \infty} 1- \sum_{j\in {\cal A}_T^c} P\Big(j \notin \bigcap_{ \lambda_n \in \Omega_2} \widehat{\cal A}_{\lambda_n} \Big )-\sum_{j_1 \in {\cal A}_T} P\Big(j_1 \notin \bigcap_{ \lambda_n \in \Omega_2} \widehat{\cal A}_{\lambda_n} \Big )\nonumber\\
&\ge&\lim_{n\rightarrow \infty} 1-\frac{p-p_0}{p}-p_0\zeta_n = \frac{p_0}{p} >0.\nonumber
\end{eqnarray}
Since $\{1,\ldots,p\} \in \bigcap_{ \lambda_n \in \Omega_2} \widehat{\cal A}_{\lambda_n}^{*b}$ implies $\sup_{ \lambda_n \in \Omega_2}\hat{s}^{*b}(\Psi,\lambda_n,m)=-1$, then
$$
\lim_{n \rightarrow \infty} E\Big (\sup_{\lambda_n \in \Omega_2} \hat{s}^{*b}(\Psi,\lambda_n,m) \Big ) \leq 1 - \lim_{n \rightarrow \infty} P\Big (\sup_{\lambda_n \in \Omega_2} \hat{s}^{*b}(\Psi,\lambda_n,m) = -1 \Big ) \leq 1- \frac{p_0}{p}.
$$
In addition, the strong law of large number for U-statistics \citep{Hoe61} implies that
\begin{eqnarray*}
B^{-1}\sum_{b=1}^B \sup_{\lambda_n \in \Omega_2}  \hat{s}^{*b}(\Psi,\lambda_n,m) \stackrel{a.s.}{\longrightarrow} E \Big( \sup_{\lambda_n \in \Omega_2} \hat{s}^{*b}(\Psi,\lambda_n,m) \Big) ~\mbox{as}~B \rightarrow \infty.
\end{eqnarray*}
Note that $\sup_{\lambda_n \in \Omega_2}\hat{s}(\Psi,\lambda_n,m) \le B^{-1}\sum_{b=1}^B \sup_{\lambda_n \in \Omega_2}  \hat{s}^{*b}(\Psi,\lambda_n,m)$, it then follows immediately that $P(\sup_{\lambda_n \in \Omega_2}\hat{s}(\Psi,\lambda_n,m) \le 1- \frac{p_0}{p}) \rightarrow 1$ and hence $P(\sup_{\lambda_n \in \Omega_2}\hat{s}(\Psi,\lambda_n,m) < 1-2 \alpha_n) \rightarrow 1$. Therefore $P( \hat \lambda_n  \in \Omega_2 ) \rightarrow 0$.

Finally, to show $P ( \hat \lambda_n  \in \Omega_3 ) \rightarrow 0$, it also suffices to show
\begin{equation}
P \Big (\sup_{\lambda_n \in \Omega_3}  \hat{s}(\Psi,\lambda_n,m) < 1 - 2\alpha_n \Big ) \rightarrow 1. \label{toshow}
\end{equation}
Assumption 2 implies that for any $j \in {\cal A}_T^c$ and some $j_1 \in {\cal A}_T^c$, when $n$ is sufficiently large,
$$
P (\cap_{\lambda_n\in \Omega_3} \{j \in \widehat{\cal A}_{\lambda_n}\} )\ge c_1(1/\epsilon)>0~~\textrm{and}~~ P (\cap_{\lambda_n\in \Omega_3} \{j_1 \notin \widehat{\cal A}_{\lambda_n}\} )\ge c_2(\tau)>0.
$$
Therefore, it follows from the independence between two sub-samples that
\begin{eqnarray*}
P \Big ( \bigcap_{\lambda_n\in \Omega_3} \{ \widehat{\cal A}_{1\lambda_n}^{*b} \neq \widehat{\cal A}_{2\lambda_n}^{*b} \} \Big ) &\geq& P \Big (\bigcap_{\lambda_n\in \Omega_3} \bigcup_{j \in {\cal A}_T^c} \{  j\notin \widehat{\cal A}_{1\lambda_n}^{*b}, j\in \widehat{\cal A}_{2\lambda_n}^{*b}\} \Big)\nonumber\\
&\geq& P\Big(  \bigcap_{\lambda_n\in \Omega_3} \{  j_1\notin \widehat{\cal A}_{1\lambda_n}^{*b}, j_1\in \widehat{\cal A}_{2\lambda_n}^{*b}\} \Big)\nonumber\\
&=& P\Big(  \bigcap_{\lambda_n\in \Omega_3} \{  j_1\notin \widehat{\cal A}_{1\lambda_n}^{*b}\} \Big)P\Big( \bigcap_{\lambda_n\in \Omega_3} \{  j_1\in \widehat{\cal A}_{2\lambda_n}^{*b}\} \Big ) \nonumber, \\
&\geq& c_1(1/\epsilon)c_2(\tau).
\end{eqnarray*}
Since the event $ \bigcap_{\lambda_n\in \Omega_3} \{ \widehat{\cal A}_{1\lambda_n}^{*b} \neq \widehat{\cal A}_{2\lambda_n}^{*b} \} $ implies that $\sup_{\lambda_n \in \Omega_3}\hat{s}^{*b}(\Psi,\lambda_n,m)\le c_3$
with $c_3=$ \linebreak[4] $\max_{{\cal A}_1\ne{\cal A}_2}\kappa({\cal A}_1, {\cal A}_2) \leq \frac{p-1}{p}$ where ${\cal A}_1, {\cal A}_2 \subset \{1,\cdots,p\}$, we have, for sufficiently large $n$,
$$
P\Big(\sup_{\lambda_n \in \Omega_3}  \hat{s}^{*b}(\Psi,\lambda_n,m)\le c_3 \Big)\ge c_1(1/\epsilon)c_2(\tau).
$$
Therefore, for sufficiently large $n$ and any $b>0$,
\begin{eqnarray*}
E \Big( \sup_{\lambda_n \in \Omega_3} \hat{s}^{*b}(\Psi,\lambda_n,m) \Big)\le  1-c_1(1/\epsilon)c_2(\tau)(1-c_3).
\end{eqnarray*}
Again, by the strong law of large number for U-statistics \citep{Hoe61} and the fact that \linebreak[4] $\sup_{\lambda_n \in \Omega_3}\hat{s}(\Psi,\lambda_n,m) \le B^{-1}\sum_{b=1}^B \sup_{\lambda_n \in \Omega_3}  \hat{s}^{*b}(\Psi,\lambda_n,m)$, we have
$$
P\Big(\sup_{\lambda_n \in \Omega_3}\hat{s}(\Psi,\lambda_n,m) \le 1-c_1(1/\epsilon)c_2(\tau)(1-c_3)\Big) \rightarrow 1.
$$
For any $\epsilon$, $c_1(1/\epsilon)c_2(\tau)(1-c_3)$ is strictly positive and $\alpha_n\rightarrow 0$, we have
$$
P\Big(\sup_{\lambda_n \in \Omega_3} \hat{s}(\Psi,\lambda_n,m)< 1-2\alpha_n\Big) \ge P\Big(\sup_{\lambda_n \in \Omega_3} \hat{s}(\Psi,\lambda_n,m)\le 1-c_1(1/\epsilon)c_2(\tau)(1-c_3)\Big) \rightarrow 1,
$$
and hence $(\ref{toshow})$ is verified and $P \Big ( \hat \lambda_n  \in \Omega_3 \Big ) \rightarrow 0$.

Combining the above results, we have for any $\epsilon>0$,
\begin{equation}
\lim_{n \rightarrow \infty} \lim_{B \rightarrow \infty} P \Big ( r_n/\epsilon \le \hat \lambda_n \le \lambda_n^* \Big ) = 1.
\label{eqn:probconv}
\end{equation}
Furthermore, since for any $\epsilon>0$,
\begin{eqnarray*}
P(\widehat{\cal A}_{\hat{\lambda}_n} = {\cal A}_T) &\ge& P\Big (  \widehat{\cal A}_{\hat{\lambda}_n}={\cal A}_T,~r_n/\epsilon \le \hat \lambda_n \le \lambda_n^* \Big )\\
&\geq& P\Big ( \bigcap_{r_n/\epsilon \le \lambda_n \le \lambda_n^*} \{\widehat{\cal A}_{\lambda_n}={\cal A}_T \} \Big ) + P\Big ( r_n/\epsilon \le \hat \lambda_n \le \lambda_n^* \Big ) -1.
\end{eqnarray*}
Therefore, the desired selection consistency directly follows from $(\ref{eqn:probconv})$ and Assumption 1 by letting $\epsilon\rightarrow 0$.  \hfill $\blacksquare$\\

\noindent{\bf Proof of Theorem \ref{nonempty2}:} We prove Theorem \ref{nonempty2} by similar approach as in the proof of Theorem 1. For any $\epsilon>0$, we denote $\Omega_1=\{\lambda_n: \lambda_n > \lambda_n^* \}$, $\Omega_2=\{\lambda_n: r_n^{-1} \lambda_n \le \tau \}$ and $\Omega_3=\{\lambda_n: \tau \leq r_n^{-1} \lambda_n \leq 1/\epsilon \}$, where $\tau$ is selected so that $\tau < 1/\epsilon$ and $p_n(1-c_{1n}(\tau)) \succ \alpha_n$. Then we just need to show that $P( \hat \lambda_n  \in \Omega_1 \cup \Omega_2 \cup \Omega_3) \rightarrow 0$. The probability $P( \hat \lambda_n  \in \Omega_1)\rightarrow 0$ for any $\epsilon>0$ can be proved similarly as in Theorem 1.

In addition, Lemma 1 implies that $P(\hat{s}(\Psi,\lambda_n^*,m)\ge 1-\alpha_n)\ge 1- 2\epsilon_n/\alpha_n$, and the definition of $\hat \lambda_n$ leads to $P(\hat{s}(\Psi,\hat{\lambda}_{n},m)\ge (1-\alpha_n)(1-\alpha_n))\ge 1-2\epsilon_n/\alpha_n$, and hence
$$
P \Big (\hat{s}(\Psi,\hat{\lambda}_{n},m) \geq 1-2\alpha_n \Big ) \ge P \Big (\hat{s}(\Psi,\hat{\lambda}_{n},m) \geq (1-\alpha_n)(1-\alpha_n) \Big ) \ge 1- \frac{2\epsilon_n}{\alpha_n}\rightarrow 1.
$$
To show $P( \hat \lambda_n  \in \Omega_2)\rightarrow 0$, it suffices to show $P( \sup_{\lambda_n \in \Omega_2} \hat{s}(\Psi,\lambda_n,m) < 1 - 2\alpha_n) \rightarrow 1$, which can be verified by slightly modifying the proof of $(\ref{toshow0})$.
Assumption 2a implies that for any $j \in {\cal A}_T^c$ and $j_1 \in {\cal A}_T$, we have
$$
P\Big(j \in \bigcap_{ \lambda_n \in \Omega_2} \widehat{\cal A}_{\lambda_n} \Big ) \ge c_{1n}(\tau) ~~\textrm{and}~~
P\Big(j_1 \in \bigcap_{ \lambda_n \in \Omega_2} \widehat{\cal A}_{\lambda_n} \Big ) \ge 1-\zeta_n.
$$
As shown in Theorem 1, it implies that
$$
P\Big(\{1,\ldots,p\} \in \bigcap_{ \lambda_n \in \Omega_2} \widehat{\cal A}_{\lambda_n} \Big ) \ge 1-(p_n-p_{0n})(1-c_{1n}(\tau))-p_{0n}\zeta_n,
$$
and hence $E(\sup_{\lambda_n \in \Omega_2} \hat{s}^{*b}(\Psi,\lambda_n,m)) \leq 1- (p_n-p_{0n})(1-c_{1n}(\tau))-p_{0n}\zeta_n$. By the strong law of large number for U-statistics,
$$
P \Big (\sup_{\lambda_n \in \Omega_2}\hat{s}(\Psi,\lambda_n,m) \le 1- (p_n-p_{0n})(1-c_{1n}(\tau))-p_{0n}\zeta_n \Big ) \rightarrow 1.
$$
Therefore, $P( \sup_{\lambda_n \in \Omega_2} \hat{s}(\Psi,\lambda_n,m) < 1 - 2\alpha_n) \rightarrow 1$ provided that $p_n(1-c_{1n}(\tau)) \succ \alpha_n$ and $p_{n}\zeta_n\rightarrow 0$.

To show $P( \hat \lambda_n  \in \Omega_3)\rightarrow 0$, it suffices to show
\begin{equation}
P \Big ( \sup_{\lambda_n \in \Omega_3} \hat{s}(\Psi,\lambda_n,m) < 1 - 2\alpha_n \Big ) \rightarrow 1. \label{toshow2_2}
\end{equation}
Here $(\ref{toshow2_2})$ follows by modifying the proof of $(\ref{toshow})$. According to $c_4 \leq (p_n-1)/p_n$, we have
$$
E \Big(\sup_{\lambda_n \in \Omega_3} \hat{s}^{*b}(\Psi,\lambda_n,m) \Big) \le 1-p_n^{-1}c_{1n}(1/\epsilon) c_{2n}(\tau).
$$
Therefore, following the same derivation as in Theorem 1, we have
$$
P\Big(\sup_{\lambda_n \in \Omega_3} \hat{s}(\Psi,\lambda_n,m)\le 1-p_n^{-1}c_{1n}(1/\epsilon)c_{2n}(\tau)\Big) \rightarrow 1.
$$
This, together with the assumptions that $\alpha_n \prec  p_n^{-1}c_{1n}(1/\epsilon)c_{2n}(\tau)$ for any $\epsilon$ and $\tau$, leads to the convergence in $(\ref{toshow2_2})$, which completes the proof of Theorem 2.  \hfill $\blacksquare$

\end{document}


\title{Consistent Selection of Tuning Parameters via Variable Selection Stability (Supplementary Material)}

\author{\name Wei Sun \email sun244@purdue.edu \\
       \addr Department of Statistics\\
       Purdue University\\
       West Lafayette, IN 47907, USA
       \AND
       \name Junhui Wang \email junhui@uic.edu \\
       \addr Department of Mathematics, Statistics, and Computer Science\\
       University of Illinois at Chicago\\
       Chicago, IL 60607, USA
       \AND
       \name Yixin Fang \email Yixin.Fang@nyumc.org \\
       \addr Departments of Population Health and Environmental Medicine \\
       New York University\\
       New York, NY 10016, USA}

\editor{Xiaotong Shen}

\maketitle

In this supplementary material, we provide Lemmas $2$ and $3$ and their proofs.

Suppose that $\bx_1,\ldots,\bx_n$ are i.i.d. from a probability distribution with mean $0$ and finite covariance matrix $C=(C_{jk})$.

{\it Assumption S1}: Assume that $\bx_1$ has finite fourth moment, that is, $E(x_{1i} x_{1j} x_{1k} x_{1l})$ is finite for any $1 \leq i,j,k,l \leq p$.\\

\noindent{\bf Lemma 2} {\it
Suppose that Assumption S1 is met. Assumptions 1 and 2 are satisfied by the lasso regression and the SCAD with $r_n=n^{-1/2}$ and $s_n =o(1)$ under the assumptions in \citet*{Zha06} or \citet*{Fan01}, and by the adaptive lasso with $r_n=n^{-1}$ and $s_n=n^{-1/2}$ under the assumptions in \citet*{Zou06}, on the random splitting subsamples generated in Algorithm 1.
}\\

\noindent{\bf Proof of Lemma 2:} First, for random variables $w_i \sim Bern(1/2)$; $i=1,\ldots,n$ that are independent with $\bx_i$'s and satisfy $\sum_{i=1}^n w_i = \lfloor n/2 \rfloor$, we show that $\frac{2}{n}\sum_{i=1}^nw_i\bx_i\bx_i^T \stackrel{p}{\rightarrow} C$.

For fixed $p$, it suffices to show the componentwise convergence in probability,
\begin{equation}
S_n=\frac{1}{n}\sum_{i=1}^n 2w_i x_{ij}x_{ik} \stackrel{p}{\longrightarrow} C_{jk}.
\label{WLLN}
\end{equation}
Note that $E(w_i)=E(w_i^2)=1/2$, and thus
$$
E(S_n)=\frac{1}{n}\sum_{i=1}^n E(2w_ix_{ij}x_{ik})=\frac{1}{n}\sum_{i=1}^n E(2w_i)E(x_{ij}x_{ik})=C_{jk},
$$
following the independence between $w_i$ and $\bx_i$.

In addition,
\begin{eqnarray}
\var(S_n) &=& E(S_n^2) - E(S_n)^2 = E\left( \Big (\frac{1}{n}\sum_{i=1}^n 2w_ix_{ij}x_{ik} \Big )^2\right ) ~-~ C_{jk}^2\nonumber\\
&=& \frac{4}{n^2} \left( E \Big (\sum_{i=1}^n w_i^2x_{ij}^2x_{ik}^2 \Big ) ~+~ \sum_{i\ne l} E(w_iw_l)E(x_{ij}x_{ik}) E(x_{lj}x_{lk}) \right) ~-~ C_{jk}^2\nonumber\\
&=& \frac{4}{n^2} E \Big (\sum_{i=1}^n w_i^2x_{ij}^2x_{ik}^2 \Big ) ~+~ \frac{4(n-1)C_{jk}^2}{n}\cov(w_1, w_2) ~-~ \frac{C_{jk}^2}{n} \nonumber\\
&\le& \frac{4}{n^2} E \Big (\sum_{i=1}^n w_i^2x_{ij}^2x_{ik}^2 \Big ) = \frac{2}{n} E(x_{ij}^2x_{ik}^2)\rightarrow 0, \nonumber
\end{eqnarray}
where the inequalities follow from the fact $\cov(w_1, w_2)<0$, $E(x_{ij}x_{ik})=E(x_{lj}x_{lk})=C_{jk}$, and $E(w_i^2)=1/2$, and the convergence is due to the finite fourth moment of $\bx_i$. The Chebyshev's inequality immediately implies that
$\frac{2}{n}\sum_{i=1}^nw_i\bx_i\bx_i^T \stackrel{p}{\rightarrow} C$.

Next we prove Lemma 2 for (i) the lasso regression, (ii) the adaptive lasso,  and (iii) the SCAD, respectively.

(i): The lasso regression. When the original assumption $\frac{1}{n}\sum_{i=1}^n\bx_i\bx_i^T \rightarrow C$ is replaced by {\it Assumption S1}, the proof follows from the above convergence in probability statement and slight modification of some existing results in literature. Specifically, the existence of $r_n$ and $s_n$ for selection consistency in Assumption 1 can be verified as in Section 2.1 of \citet*{Zha06}. The condition $(3)$ in Assumption 1 is a direct result from Assumption 2, which will be shown after we verify conditions in Assumption 2 based on Lemma C.2 in \citet*{Bac09}.

Denote the permutated subsample $(w_i\bx_1,\ldots,w_n\bx_n)$ as $\bold{Z}=(\bz_1,\ldots,\bz_{m})^T$ with $m=\lfloor n/2 \rfloor$. Denote $Q=\frac{1}{m} \bold{Z}^T\bold{Z} = \frac{1}{m}\sum_{i=1}^{m}\bz_i\bz_i^T$, $\lambda_{\min}(Q)$ as the minimal eigenvalue of $Q$, $q=\bold{Z}^T\epsilon/m$, the true coefficient as $\beta^*$, and a sign pattern $s=\{1,0,-1\}^p$ such that for any $j\in \{1,\ldots,p\}$, $s_j=\textrm{sign}(\beta_j)$. For simplicity, we denote $J=\widehat{\cal A}_{\lambda_m}$, $\bold{J}={\cal A}_T$, and $s_J$ as the sign pattern of variables indexed by $J$. Let $M(\beta)=\min_{j\in \{1,\ldots,p\}, \beta_j\neq 0}|\beta_j|$ as the smallest magnitude of non-zero elements in $\beta$, and $\|C\|_{\infty}$ as the largest magnitude of all the elements in matrix $C$. According to Lemma C.2 in \citet*{Bac09}, when the selected active set is over-fitting such that $s_{\bold{J}}=\textrm{sign}(\beta_{\bold{J}})$ and $J\supset \bold{J}$, we have that $s$ is selected if and only if
\begin{eqnarray}
&&\|Q_{J^c,J}Q_{J,J}^{-1}q_J-q_{J^c}-\lambda_mQ_{J^c,J}Q_{J,J}^{-1}s_J\|_{\infty} \le \lambda_m; \label{C3}\\
&&\textrm{sign}\Big( \beta^*_{\bold{J}} +  (Q_{J^c,J}Q_{J,J}^{-1}q_J - \lambda_m Q_{J,J}^{-1}s_J)_{\bold{J}} \Big)=\textrm{sign}(\beta^*_{\bold{J}});\label{C4}\\
&&\textrm{sign}\Big(Q_{J,J}^{-1}q_J-\lambda_mQ_{J,J}^{-1}s_J \Big)_{J\cap \bold{J}^c}=s_{J\cap \bold{J}^c}.\label{C5}
\end{eqnarray}
Therefore, for a particular over-fitting sign pattern $\tilde{s}$ with $j$th noise variable selected in the active set $J$, we have $\{j\in J\}\supseteq \{\textrm{sign}(\hat{\beta}_n) = \tilde{s}\}$, where $\{\textrm{sign}(\hat{\beta}_n) = \tilde{s}\}$ is equivalent to the conditions of $\eqref{C3}$-$\eqref{C5}$ with $s=\tilde{s}$. For short, we denote $\{\eqref{C4}^c\}$ as the complement of condition in $\eqref{C4}$, and $\{\eqref{C3},\eqref{C5}\}^c$ as the complement of conditions in $\eqref{C3}$ and $\eqref{C5}$, respectively. When $\sqrt{m}\lambda_m \le \lambda_0\in (0,\infty)$, Proposition 2.4 in \citet*{Bac09} leads to
\begin{eqnarray}
&& \bigcup_{\lambda_m: \sqrt{m}\lambda_m \le \lambda_0}\left\{\eqref{C4}^c\right\} \nonumber\\
&\subseteq& \bigcup_{\lambda_m: \sqrt{m}\lambda_m \le \lambda_0}\left\{ \sqrt{m}\lambda_m > \frac{\sqrt{m}M(\beta^*)\lambda_{\min}(Q)}{2\sqrt{p}}, ~\textrm{or}~ \|(Q_{J,J}^{-1}q_J)_{\bold{J}}\|_2 > \frac{M(\beta^*)}{2} \right\}\nonumber\\
&\subseteq& \left\{\lambda_0 > \frac{\sqrt{m}M(\beta^*)\lambda_{\min}(Q)}{2\sqrt{p}}, ~\textrm{or}~ \|(Q_{J,J}^{-1}q_J)_{\bold{J}}\|_2 > \frac{M(\beta^*)}{2} \right\},\label{bound1}
\end{eqnarray}
with the right hand side in $\eqref{bound1}$ having probability tending to $0$, and as $m\rightarrow \infty$
\begin{equation}
\bigcup_{\lambda_m: \sqrt{m}\lambda_m \le \lambda_0}\left\{\eqref{C3}, \eqref{C5}\right\}^c \rightarrow \{v\notin {\cal C}(\tilde{s},\lambda_0)\},\label{bound2}
\end{equation}
where $v$ is normal with zero mean and covariance matrix $Q$, and ${\cal C}(\tilde{s},\lambda_0)$ is a convex set and its complement also have non-empty interior, and hence $P(v\notin {\cal C}(\tilde{s},\lambda_0))$ is strictly within $(0,1)$ for any fixed $\lambda_0$. Therefore, as $m\rightarrow \infty$, combining $\eqref{bound1}$ and $\eqref{bound2}$ leads to
\begin{eqnarray*}
&&P\Big(\bigcap_{\lambda_m: \sqrt{m}\lambda_m \le \lambda_0} \{j \in \widehat{\cal A}_{\lambda_m}\}\Big) \\
&\ge& P\Big(\bigcap_{\lambda_m: \sqrt{m}\lambda_m \le \lambda_0} \{\textrm{sign}(\hat{\beta}_n) = \tilde{s}\}\Big)\\
&=& P\Big(\bigcap_{\lambda_m: \sqrt{m}\lambda_m \le \lambda_0}\{\eqref{C3}, \eqref{C5}\}\cap \{\eqref{C4}\}\Big)\\
&\ge& 1 -  P\Big(\bigcup_{\lambda_m: \sqrt{m}\lambda_m \le \lambda_0}\{\eqref{C3}, \eqref{C5}\}^c \Big) - P\Big(\bigcup_{\lambda_m: \sqrt{m}\lambda_m \le \lambda_0} \{\eqref{C4}\}^c\Big)\\
&\rightarrow& 1 - P(v\notin {\cal C}(\tilde{s},\lambda_0))\in (0,1),
\end{eqnarray*}
and hence condition $(5)$ in Assumption 2 is verified. In addition, as shown in Proposition 1 in \citet*{Bac08}, when $\lambda_0$ converges to $0$, $P(v\notin {\cal C}(\tilde{s},\lambda_0))\rightarrow 0$, and hence $c_1(\lambda_0)=1 - P(v\notin {\cal C}(\tilde{s},\lambda_0))\rightarrow 1$. The condition $(6)$ in Assumption 2 can be proved by defining the particular sign pattern $\tilde{s}$ to be the one with $j$th noise variable not selected in the active set $J$, then $\{j\notin J\}\supseteq\{\textrm{sign}(\hat{\beta}_n) = \tilde{s}\}$. All the proof can be derived following similar approach as above. Therefore, for any $j\in{\cal A}_T^c$, we have $P(\cap_{r_n^{-1} \lambda_n \geq \lambda_0} \{j \notin \widehat{\cal A}_{\lambda_n} \} )\ge c_2(\lambda_0)$ with $c_2(\lambda_0)\rightarrow 1$ as $\lambda_0\rightarrow \infty$. In addition, after a slight modification of Proposition 2.5 in \citet*{Bac09}, we can show that uniformly over $\lambda_m$ such that $\sqrt{m}\lambda_m \le \lambda_0$, all the important variable will be selected with probability tending to $1$, which verifies condition $(4)$ in Assumption 2. This ends the verification of Assumption 2 for the lasso regression.

Finally we show condition (3) in Assumption 1 for the lasso regression. Note that
\begin{eqnarray}
&&\bigcap_{\lambda_0m^{-1/2} \le \lambda_m \le \lambda_m^*}\Big\{\widehat{\cal A}_{\lambda_m} = {\cal A}_T \Big\}  \nonumber\\
&=& \bigcap_{\lambda_0m^{-1/2} \le \lambda_m \le \lambda_m^*} \left\{ \bigcap_{j\in {\cal A}_T}\{j\in \widehat{\cal A}_{\lambda_m} \}\right\} \bigcap \left\{\bigcap_{j_1\in {\cal A}_T^c}\{j_1\notin \widehat{\cal A}_{\lambda_m} \}  \right\}\nonumber\\
&\supset& \Big\{\bigcap_{\lambda_m \le \lambda_m^*; j\in {\cal A}_T} \{j\in \widehat{\cal A}_{\lambda_m} \}\Big\} \bigcap \Big\{\bigcap_{\lambda_m \ge \lambda_0m^{-1/2}; j_1\in {\cal A}_T^c} \{j_1 \notin \widehat{\cal A}_{\lambda_m} \}\Big\}. \nonumber
\end{eqnarray}
Following the similar strategy in the proof of conditions $(4)$ and $(6)$, the selection consistency in \citet*{Zha06} and Proposition 2.5 in \citet*{Bac09} imply that all the important variables will be included uniformly over $\lambda_m \le \lambda_m^*$, and all the noisy variables will be excluded in the active set $\widehat{\cal A}_{\lambda_m}$ uniformly over $\lambda_m \ge \lambda_0m^{-1/2}$. Therefore, when $n$ is sufficiently large,
\begin{eqnarray}
&&P\Big(\bigcap_{\lambda_0m^{-1/2} \le \lambda_m \le \lambda_m^*}\Big\{\widehat{\cal A}_{\lambda_m} = {\cal A}_T \Big\}\Big) \nonumber\\
&\ge& P\Big( \bigcap_{\lambda_m \le \lambda_m^*; j\in {\cal A}_T} \{j\in \widehat{\cal A}_{\lambda_m} \}\Big) + P\Big( \bigcap_{\lambda_0m^{-1/2} \le \lambda_m; j_1\in {\cal A}_T^c} \{j_1 \notin \widehat{\cal A}_{\lambda_m} \}\Big) -1\nonumber\\
&\ge& c_2(\lambda_0)-\zeta_n.\nonumber
\end{eqnarray}
Since $\zeta_n\rightarrow 0$ and $\lim_{\lambda_0\rightarrow \infty}c_2(\lambda_0) = 1$ as shown above, letting $c_0(\lambda_0)=1-c_2(\lambda_0)/2$ leads to $(3)$ in Assumption 1. This ends the verification for lasso regression.

(ii): The adaptive lasso. When the original assumption $\frac{1}{n}\sum_{i=1}^n\bx_i\bx_i^T \rightarrow C$ is replaced by {\it Assumption S1}, the selection consistency established in \citet*{Zou06} when $n\lambda_n \rightarrow \infty$ and $\sqrt{n}\lambda_n \rightarrow 0$ is still valid with the above convergence in probability statement. In specific, we also denote the permutated subsample $(w_i\bx_1,\ldots,w_n\bx_n)$ as $\bold{Z}=(\bz_1,\ldots,\bz_{m})^T$ with $m=\lfloor n/2 \rfloor$. It is shown above that $\frac{1}{m} \bold{Z}^T\bold{Z} = \frac{1}{m}\sum_{i=1}^{m}\bz_i\bz_i^T \stackrel{p}{\rightarrow} C$. Denote $\beta^*$ as the true coefficient, $\beta=\beta^*+\frac{u}{\sqrt{m}}$, and
$$
\Psi_m(u)=\Big\|\bold{y}-\sum_{j=1}^p \bold{z}_{(j)}\Big(\beta_j^*+\frac{u_j}{\sqrt{m}}\Big)\Big\|^2+
m\lambda_m\sum_{j=1}^{p}\frac{\Big|\beta_j^*+\frac{u_j}{\sqrt{m}}\Big|}{|\hat{\beta}_j^{L}|},
$$
where $\hat{\beta}_j^{L}$ is the estimator from the lasso regression. Let $\hat{u}_m=\arg \min \Psi_m(u)$, $\hat \beta_m=\beta^*+\frac{\hat{u}_m}{\sqrt{m}}$, and  $V_m(u)=\Psi_m(u)-\Psi_m(0)$ with
$$
V_m(u)=u^T\Big(\frac{\bold{Z}^T\bold{Z}}{m}\Big)u-\frac{2\epsilon^T\bold{Z}}{\sqrt m}u + \sqrt{m}\lambda_m \sum_{j=1}^{p} \frac{\sqrt{m}\Big(\Big|\beta_j^*+\frac{u_j}{\sqrt{m}}\Big|
-|\beta_j^*|\Big)}{|\hat{\beta}_j^{L}|}.
$$
Note that $\frac{\bold{Z}^T\bold{Z}}{m} \stackrel{p}{\rightarrow} \bold{C}$, $\frac{\epsilon^T\bold{Z}}{\sqrt m} \stackrel{d}{\rightarrow} W^T \sim N(0, \sigma^2\bold{C})$ from the central limit theorem. Similar to the fixed design case in \citet*{Zou06}, we can show that with probability tending to 1, the asymptotic normality of $\hat{u}_m$ holds on ${\cal A}_T$ and $\hat{u}_m\rightarrow 0$ on ${\cal A}_T^c$. In addition, for any $j \notin {\cal A}_T$, it's sufficient to show $P(j \in \widehat{\cal A}_{\lambda_m})\rightarrow 0$. Note that when $j \in \widehat{\cal A}_{\lambda_m}$, the Karush-Kuhn-Tucker (KKT) conditions imply that $2\bold{z}_{(j)}^T(\bold{y}-\bold{Z}\hat{\beta}_n) = m\frac{\lambda_m}{|\hat{\beta}_j^{L}|}$, where $\sqrt{m}\frac{\lambda_m}{|\hat{\beta}_j^{L}|} \stackrel{p}\rightarrow \infty$, and
\begin{equation}
\frac{2\bold{z}_{(j)}^T(\bold{y}-\bold{Z}\hat{\beta}_m)}{\sqrt{m}}=\frac{\bold{z}_{(j)}^T\bold{Z} \sqrt{m}(\beta^*-\hat{\beta}_m)}{m} + \frac{2\bold{z}_{(j)}^T\epsilon}{\sqrt{m}}.
\nonumber
\end{equation}
Note that $\frac{\bold{Z}^T\bold{Z}}{m} \stackrel{p}{\rightarrow} \bold{C}$ and $\sqrt{m}(\beta^*-\hat{\beta}_m)$ is asymptotic normal as shown above, Slutsky's theorem implies that both $2\bold{z}_{(j)}^T\bold{Z} \sqrt{m}(\beta^*-\hat{\beta}_m)/m$ and $2\bold{z}_{(j)}^T\epsilon/\sqrt{m}$ converge in distribution to normal. Therefore, $P(j \in \widehat{\cal A}_{\lambda_m})\rightarrow 0$.

Next we verify Assumption 2 for the permutated subsample $\bold{Z}=(\bz_1,\ldots,\bz_{m})^T$. When $\lambda_m \preceq m^{-1}$, we have $\lambda_m \prec m^{-1/2}$, and hence the asymptotic normality of $\hat{\beta}_m$ still holds for any satisfied $\lambda_m$ \citep{Zou06}. This implies condition $(4)$ in Assumption 2 directly. It then suffices to consider the event $j \notin \widehat{\cal A}_{\lambda_m}$ for any $j \in {\cal A}_T^c$. Note that when $j \notin \widehat{\cal A}_{\lambda_m}$, the Karush-Kuhn-Tucker (KKT) conditions imply that
\begin{equation}
\Big|2\bold{z}_{(j)}^T(\bold{y}-\bold{Z}\hat{\beta}_n)\Big| \le m\frac{\lambda_m}{|\hat{\beta}_j^{L}|}.
\nonumber
\end{equation}
In addition,
\begin{equation}
\frac{2\bold{z}_{(j)}^T(\bold{y}-\bold{Z}\hat{\beta}_m)}{\sqrt{m}}=\frac{2\bold{z}_{(j)}^T\bold{Z} \sqrt{m}(\beta^*-\hat{\beta}_m)}{m} + \frac{2\bold{z}_{(j)}^T\epsilon}{\sqrt{m}}.
\nonumber
\end{equation}
By the asymptotic normality of $\hat{\beta}_m$ and $\frac{\bold{Z}^T\bold{Z}}{m} \stackrel{p}{\rightarrow} \bold{C}$, the Slutsky's theorem implies that
$2\bold{z}_{(j)}^T\bold{Z} \sqrt{m}(\beta^*-\hat{\beta}_m)/m \stackrel{d}{\rightarrow} N(0, \Delta_1)$ for some $\Delta_1$, and $2\bold{z}_{(j)}^T\epsilon/\sqrt{m} \stackrel{d}{\rightarrow} N(0, \Delta_2)$ for some $\Delta_2$. In addition, when $m\lambda_m \le \lambda_0$ for some $\lambda_0\in (0,\infty)$, we have
\begin{eqnarray}
\left\{\bigcup_{\lambda_m: m\lambda_m \le \lambda_0} \{j \notin \widehat{\cal A}_{\lambda_m}\} \right\} &\subseteq& \left\{\bigcup_{\lambda_m: m\lambda_m \le \lambda_0} \Big\{\Big|2\bold{z}_{(j)}^T(\bold{y}-\bold{Z}\hat{\beta}_m)\Big| \le \frac{m\lambda_m}{|\hat{\beta}_j^{L}|}\Big\} \right\}\nonumber\\
&\subseteq& \Big\{\Big|2\bold{z}_{(j)}^T(\bold{y}-\bold{Z}\hat{\beta}_m)\Big| \le \frac{\lambda_0}{|\hat{\beta}_j^{L}|} \Big\}.\nonumber
\end{eqnarray}
Therefore,
\begin{eqnarray*}
P\Big(\bigcup_{\lambda_m: m\lambda_m \le \lambda_0} \{j \notin \widehat{\cal A}_{\lambda_m}\}\Big)
&\le& P\Big(\Big|2\bold{z}_{(j)}^T(\bold{y}-\bold{Z}\hat{\beta}_m)\Big| \le \frac{\lambda_0}{|\hat{\beta}_j^{L}|}\Big)\\
&=& P\Big(\Big|\frac{2\bold{z}_{(j)}^T\bold{Z} \sqrt{m}(\beta^*-\hat{\beta}_m)}{m} + \frac{2\bold{z}_{(j)}^T\epsilon}{\sqrt{m}}\Big|\le \frac{\lambda_0}{|\sqrt{m}\hat{\beta}_j^{L}|}\Big).
\end{eqnarray*}
The DCT theorem implies that for sufficiently large $m$,
$$
P\Big(\bigcup_{\lambda_m: m\lambda_m \le \lambda_0} \{j \notin \widehat{\cal A}_{\lambda_m}\}\Big)\le 1 - c_0(\lambda_0),
$$
with $c_1(\lambda_0)$ being a strictly positive constant in $(0,1)$, and hence
$$
P\Big( \bigcap_{\lambda_m: m\lambda_m \le \lambda_0} \{j \in \widehat{\cal A}_{\lambda_n}\} \Big)\ge c_0(\lambda_0).
$$

In addition, the condition $(6)$ in Assumption 2 can be verified by slightly modifying the proof in the lasso regression. Specifically, we define $s$ to be the one with $j$th variable not selected in the active set $J$. Then we only need to replace $\lambda_m$ with $\lambda_m/\widehat{\beta}^L_j$ for $j\in J$ in $\eqref{bound1}$ and $\eqref{bound2}$. When $m\lambda_m \le \lambda_0$, we have that $\eqref{bound1}$ is replaced with
\begin{eqnarray}
&&\bigcup_{\lambda_m: m\lambda_m \le \lambda_0}\left\{\eqref{C4}\right\}^c \nonumber\\
&\subseteq& \bigcup_{\lambda_m: m\lambda_m \le \lambda_0}\left\{ \frac{m\lambda_m}{M(\sqrt{m}\widehat{\beta}^L_J)} > \frac{\sqrt{m}M(\beta^*)\lambda_{\min}(Q)}{2\sqrt{p}}, ~\textrm{or}~ \|(Q_{J,J}^{-1}q_J)_{\bold{J}}\|_2 > \frac{M(\beta^*)}{2} \right\}\nonumber\\
&\subseteq& \left\{ \lambda_0 > \frac{\sqrt{m}M(\beta^*)\lambda_{\min}(Q)M(\sqrt{m}\widehat{\beta}^L_J)}{2\sqrt{p}}, ~\textrm{or}~ \|(Q_{J,J}^{-1}q_J)_{\bold{J}}\|_2 > \frac{M(\beta^*)}{2} \right\},\label{bound3}
\end{eqnarray}
with the right hand side in $\eqref{bound3}$ still having probability tending to $0$ since $\sqrt{m}\widehat{\beta}^L_j=O_p(1)$ for any $j\in J$. In addition, $\eqref{bound2}$ is replaced with
$$
\bigcup_{\lambda_m: m\lambda_m \le \lambda_0}\left\{\eqref{C3}, \eqref{C5}\right\}^c \rightarrow \{v\notin {\cal C}(s,\lambda_0)\}.
$$
Therefore, we still have
$$
P\Big( \bigcap_{\lambda_m: m\lambda_m \le \lambda_0} \{j \notin \widehat{\cal A}_{\lambda_m}\} \Big)\ge c_1(\lambda_0).
$$
This ends the proof of Assumption 2 for the adaptive lasso.

(iii): The SCAD. \citet*{Fan01} showed that the SCAD is selection consistent under the random design when $\sqrt{m}\lambda_m \rightarrow \infty$ and $\lambda_m \rightarrow 0$. In addition, condition $(3)$ in Assumption 1 follows after the verification of Assumption 2 by similar approach as in the proof of lasso regression case.

Next, we show Assumption 2 for SCAD. It then suffices to consider the event $j \notin \widehat{\cal A}_{\lambda_m}$ for any $j \in {\cal A}_T^c$. In fact, the SCAD minimizes
\begin{eqnarray*}
Q(\beta)= \Big\|\bold{y}-\sum_{j=1}^p \bold{z}_{(j)}\beta_j\Big\|^2+ m\sum_{j=1}^{p}{p_{\lambda_m}(|\beta_j|)},
\end{eqnarray*}
where the penalty term satisfies
$p'_{\lambda}(\theta)=\lambda\big (I(\theta\le\lambda)+\frac{(\gamma\lambda-\theta)_{+}}{(\gamma-1)\lambda}I(\theta>\lambda)\big )$ for some $\gamma>2$ and $\theta>0$. For any $\beta \in \{ \beta: \beta - \hat{\beta}_m = O_P(m^{-1/2}) \} $, then
\begin{eqnarray*}
\frac{\partial Q(\beta)}{\partial \beta_j}&=& -2\bold{z}_{(j)}^T(\bold{y}-\bold{Z}\beta) + m p_{\lambda_m}^{'}(|\beta_j|)\mbox{sgn}(\beta_j)\\
&=& -m\lambda_m \left ( \frac{\frac{2\bold{z}_{(j)}^T\bold{Z} \sqrt{m}(\beta^*-\beta)}{m} + \frac{2\bold{z}_{(j)}^T\epsilon}{\sqrt{m}}}{\sqrt m \lambda_m} - \frac{p_{\lambda_m}^{'}(|\beta_j|)\mbox{sgn}(\beta_j)}{\lambda_m}  \right ),
\end{eqnarray*}
where $\frac{\bold{Z}^T\bold{Z}}{m} \stackrel{p}{\rightarrow} \bold{C}$, $\|\sqrt{m}(\beta^*-\beta)\| \leq \| \sqrt{m} (\beta^*-\hat{\beta}_m) \| + \| \sqrt{m} (\hat{\beta}_m-\beta) \|$ is bounded in probability due to that fact that $\hat{\beta}_m$ is a $\sqrt{m}$-consistent estimate of $\beta^*$ by Theorem 1 of \citet*{Fan01}, and $2\bold{z}_{(j)}^T\epsilon/\sqrt{m} \stackrel{d}{\rightarrow} N(0, \Delta_3)$ for some $\Delta_3$. Here condition $(4)$ can be verified by similar approach as in lasso regression case since we have the asymptotic normality of $\hat{\beta}_m$. In addition, $p_{\lambda_m}^{'}(|\beta_j|)/\lambda_m=I(\theta\le\lambda_m)+\frac{(\gamma\lambda_m-\theta)_{+}}{(\gamma-1)\lambda_m}I(\theta>\lambda_m)\le 1$. Therefore, we have
\begin{eqnarray*}
&&\bigcap_{\lambda_m: \sqrt{m}\lambda_m \le \lambda_0}\left\{ \left| \frac{\frac{2\bold{z}_{(j)}^T\bold{Z} \sqrt{m}(\beta^*-\beta)}{m} + \frac{2\bold{z}_{(j)}^T\epsilon}{\sqrt{m}}}{\sqrt m \lambda_m}\right|>
\left|\frac{p_{\lambda_m}^{'}(|\beta_j|)\mbox{sgn}(\beta_j)}{\lambda_m}\right| \right\}\\
&=&\bigcap_{\lambda_m: \sqrt{m}\lambda_m \le \lambda_0}\left\{ \left| \frac{2\bold{z}_{(j)}^T\bold{Z} \sqrt{m}(\beta^*-\beta)}{m} + \frac{2\bold{z}_{(j)}^T\epsilon}{\sqrt{m}}\right |>\sqrt m \lambda_m\frac{p_{\lambda_m}^{'}(|\beta_j|)}{\lambda_m} \right\}\\
&\supseteq& \left\{ \left| \frac{2\bold{z}_{(j)}^T\bold{Z} \sqrt{m}(\beta^*-\beta)}{m} + \frac{2\bold{z}_{(j)}^T\epsilon}{\sqrt{m}}\right |>\lambda_0 \right \},
\end{eqnarray*}
and hence
\begin{eqnarray*}
&&P\left (\bigcap_{\lambda_m: \sqrt{m}\lambda_m \le \lambda_0} \left\{\left| \frac{\frac{2\bold{z}_{(j)}^T\bold{Z} \sqrt{m}(\beta^*-\beta)}{m} + \frac{2\bold{z}_{(j)}^T\epsilon}{\sqrt{m}}}{\sqrt m \lambda_m}\right |>
\left |\frac{p_{\lambda_m}^{'}(|\beta_j|)\mbox{sgn}(\beta_j)}{\lambda_m}\right | \right\}\right )\\
&\ge&P\left ( \left| \frac{2\bold{z}_{(j)}^T\bold{Z} \sqrt{m}(\beta^*-\beta)}{m} + \frac{2\bold{z}_{(j)}^T\epsilon}{\sqrt{m}}\right |>\lambda_0 \right ).
\end{eqnarray*}
Therefore, there exists a positive probability $c_2(\lambda_0)\in (0,1)$, uniformly on $\lambda_m$,
\begin{eqnarray*}
&&\frac{\partial Q(\beta)}{\partial \beta_j}<0 ~\mbox{when}~ 0<\beta_j< Mm^{-1/2};\\
&&\frac{\partial Q(\beta)}{\partial \beta_j}>0 ~\mbox{when}~ -Mm^{-1/2}<\beta_j<0,
\end{eqnarray*}
with $M$ sufficient large such that $P\big(\sup_{\|u\|=M} Q\big(\beta^*+(m^{-1/2}+a_m)u\big)>Q(\beta^*)\big)\rightarrow 1$ and $a_m=\max\{p_{\lambda_m}^{'}(|\beta_j^*|): \beta_j^*\ne 0\}$,
which implies that for sufficiently large $m$ $P(\cap_{\lambda_m: \sqrt{m}\lambda_m \le \lambda_0}\{\hat{\beta}_j\ne 0\}) \geq c_3(\lambda_0)$ with $c_3(\lambda_0)$ strictly positive for fixed $\lambda_0$, and $c_3(\lambda_0)$ converges to $1$ as $\lambda_0\rightarrow 0$. Therefore, condition $(5)$ in Assumption 2 is verified. By similar approach, we can replace $\sqrt{m}\lambda_m \le \lambda_0$ with $\sqrt{m}\lambda_m \ge \tilde \lambda_0$, and bound the probability of the event $|2\bold{z}_{(j)}^T\bold{Z} \sqrt{m}(\beta^*-\beta)/m + 2\bold{z}_{(j)}^T\epsilon/\sqrt{m} |>\tilde \lambda_0$. Then condition $(6)$ in Assumption 2 can be verified. Therefore, Assumptions 2 is satisfied by the SCAD with $r_m=m^{-1/2}$ and $s_m=o(1)$. \hfill $\blacksquare$\\

\noindent{\bf Remark:} The convergence in $\eqref{WLLN}$ is valid under the fixed design with Assumption $S2$.

{\it Assumption S2}: Assume that $\frac{1}{n}\sum_{i=1}^n\bx_i\bx_i^T\rightarrow C$ with $C$ positive definite, and $\frac{1}{n}\sum_{i=1}^n x_{ij}^2x_{ik}^2$ is finite for any $1 \leq j,k \leq p$.\\

\noindent{\bf Proof of Remark:} For random variable $w_i$ as defined above, we can also show that
\begin{equation}
S_n=\frac{1}{n}\sum_{i=1}^n 2w_i\bx_{ij}\bx_{ik} \stackrel{p}{\longrightarrow} C_{jk}.
\label{p1}
\end{equation}
Note that $E(S_n)=\frac{1}{n}\sum_{i=1}^n 2E(w_i)\bx_{ij}\bx_{ik}=\frac{1}{n}\sum_{i=1}^n \bx_{ij}\bx_{ik}\rightarrow C_{jk}$ and
\begin{eqnarray*}
&&\var(S_n) = E\Big( \Big (\frac{1}{n}\sum_{i=1}^n 2w_ix_{ij}x_{ik} \Big )^2\Big) - C_{jk}^2 \\
&=& \frac{4}{n^2} \Big (\sum_{i=1}^n E(w_i^2)x_{ij}^2x_{ik}^2 + \sum_{i\ne l} E(w_iw_l)x_{ij}x_{ik}x_{lj}x_{lk} \Big ) - C_{jk}^2 \\
&=& \frac{4}{n^2} \Big ( \frac{1}{2} \sum_{i=1}^n x_{ij}^2x_{ik}^2 + \sum_{i\ne l} \cov(w_iw_l)x_{ij}x_{ik}x_{lj}x_{lk} \Big ) + \Big (\frac{1}{n^2} \sum_{i\ne l} x_{ij}x_{ik}x_{lj}x_{lk} - C_{jk}^2 \Big ) \\ &\rightarrow& 0,
\end{eqnarray*}
following from Assumption $S2$ and the fact that $\cov(w_i, w_l)< 0$ for $i\ne l$, $\sum_{i\ne l}\frac{x_{ij}x_{ik}}{n}\frac{x_{lj}x_{lk}}{n} \rightarrow C_{jk}^2 \ge 0$. Then the Chebyshev's inequality implies that
$$
P\Big(\Big|S_n-\frac{1}{n}\sum_{i=1}^n \bx_{ij}\bx_{ik}\Big|\ge \epsilon\Big)\le \frac{\var(S_n)}{\epsilon^2} \rightarrow 0, \textrm{~as~} n\rightarrow \infty.
$$
This together with the fact $\frac{1}{n}\sum_{i=1}^n \bx_{ij}\bx_{ik}\rightarrow C_{jk}$ imply $\eqref{p1}$. \hfill $\blacksquare$\\

\noindent{\bf Lemma 3} {\it
Suppose that Assumption S1 is met and $p_n=o(n^{1/4})$. Assumptions 1a and 2a are satisfied by the lasso regression with $r_n=n^{-1/2}$ and $s_n =o(1)$ under the assumptions (A1)-(A4) in \citet*{Bac09} on the random splitting subsamples generated in Algorithm 1.
}\\

\noindent{\bf Proof of Lemma 3:} According to the similar techniques in the proof of Lemma 2, we only need to validate conditions $(8)$-$(10)$ in Assumption 2a, then condition $(7)$ in Assumption 1a is a direct result from Assumption 2a as shown in Lemma 2.

For a particular sign pattern $\tilde{s}$ with $j$th noise variable selected in the active set $J$, we have $\{j\in J\}\supseteq \{\textrm{sign}(\hat{\beta}_n) = \tilde{s}\}$, where $\{\textrm{sign}(\hat{\beta}_n) = \tilde{s}\}$ is equivalent to the conditions $\eqref{C3}$-$\eqref{C5}$ with $s=\tilde{s}$. We first show the probability of $\eqref{bound1}$ also tends to zero for diverging $p_n$.  According to Proposition 2.4 in \citet*{Bac09}, Berry-Esseen inequality implies that
$$
P(S3)\ge 1-2p_ne^{-C_1\frac{n}{p_n}},
$$
where $C_1$ is a positive constant independent of $n$ and $p_n$. Therefore, from $\eqref{bound1}$,
$$
P\Big( \|(Q_{J,J}^{-1}q_J)_{\bold{J}}\|_2 > \frac{M(\beta^*)}{2} \Big)\le 2p_ne^{-C_1\frac{n}{p_n}}.
$$
In addition, $\sqrt{n}M(\beta^*)\lambda_{\min}(Q)/(2\sqrt{p_n})$ is unbounded and hence the right hand side in $\eqref{bound1}$ having probability tending to $0$ when $p_n=o(n)$. Next, we can show that $\eqref{bound2}$ is also valid for diverging $p_n$ based on Proposition 2.4 in \citet*{Bac09}. Specifically,
\begin{equation}
P\left(\bigcup_{\lambda_n: \sqrt{n}\lambda_n \le \lambda_0}\left\{\eqref{C3}, \eqref{C5}\right\}^c \right)- P\left(\{v\notin {\cal C}(\tilde{s},\lambda_0)\} \right) \le C_2\frac{p_n^2}{n^{1/2}},\label{bound2_lemma3}
\end{equation}
where $C_2$ is a constant independent of $n$ and $p_n$, and ${\cal C}(\tilde{s},\lambda_0)$ is defined in $\eqref{bound2}$. Here $P\left(\{v\notin {\cal C}(\tilde{s},\lambda_0)\} \right)$ is strictly within $(0,1)$ for any fixed $\lambda_0$, and when $\lambda_0$ converges to $0$, $P(v\notin {\cal C}(\tilde{s},\lambda_0))\rightarrow 0$. Therefore, for $p_n=o(n^{1/4})$ and $j\in {\cal A}_T^c$, as $n\rightarrow \infty$, combining $\eqref{bound1}$ and $\eqref{bound2_lemma3}$ leads to
$$
P\Big(\bigcap_{\lambda_n: \sqrt{n}\lambda_n \le \lambda_0} \{j \in \widehat{\cal A}_{\lambda_n}\}\Big) \rightarrow 1 - P(v\notin {\cal C}(\tilde{s},\lambda_0)),
$$
and hence condition $(9)$ in Assumption 2a is verified. The condition $(10)$ in Assumption 2a can be proved by defining the particular sign pattern $\tilde{s}$ to be the one with $j$th noise variable not selected in the active set $J$, then $\{j\notin J\}\supseteq \{\textrm{sign}(\hat{\beta}_n) = \tilde{s}\}$. Then all the proof can be derived following similar approach. In addition, Proposition 2.5 in \citet*{Bac09} implies that uniformly over $\lambda_n$ with $\sqrt{n}\lambda_n \le \lambda_0$, all the important variable will be selected with probability tending to $1$, which verifies $(8)$ in Assumption 2a. Specifically, for any $j\in {\cal A}_T$,
$$
P\Big(\bigcap_{\sqrt{n}\lambda_n \leq \lambda_0} \{j \in \widehat{\cal A}_{\lambda_n}\} \Big ) \ge 1-2p_ne^{-C_3\frac{n}{p_n}},
$$
where $C_3$ is a positive constant independent of $n$ and $p_n$. Therefore, let $\zeta_n=2p_ne^{-C_3\frac{n}{p_n}}$ satisfying condition $(8)$. This ends the verification of Assumption 2a for the lasso regression. Finally, the condition $(7)$ in Assumption 1a is also a direct result from Assumption 2a as shown in Lemma 2. This ends the verification for the lasso regression in Lemma 3. \hfill $\blacksquare$